\documentclass[journal]{IEEEtran}
\usepackage{amsmath,amsfonts}
\usepackage{algorithmic}
\usepackage{algorithm}
\usepackage{array}
\usepackage[caption=false,font=normalsize,labelfont=sf,textfont=sf]{subfig}
\usepackage{textcomp}
\usepackage{url}
\usepackage{verbatim}
\usepackage{graphicx}
\usepackage{cite}

\usepackage{xfrac}
\usepackage{caption}
\usepackage[LGR,T1]{fontenc}
\usepackage[bookmarks=false]{hyperref} 
\usepackage{hyperref}
\usepackage[dvipsnames]{xcolor}
\usepackage{pifont}       % \ding{xx}
\usepackage{bbding}
\usepackage{booktabs}
\usepackage{multicol}
\usepackage{multirow}
\usepackage{color}
\usepackage{colortbl}
\usepackage{tabularx} 
\usepackage{siunitx} % 用于表格对齐数值格式
\usepackage[table]{xcolor} % 用于表格颜色
\usepackage{caption} % 控制 caption 样式

\usepackage{caption}
\usepackage{overpic}

\definecolor{yzybest}{rgb}{0.678, 0.847, 0.902} % 蓝色
\definecolor{yzysecond}{rgb}{0.678, 0.902, 0.678} % 绿色
\definecolor{yzythird}{rgb}{0.902, 0.678, 0.678} % 红色

\usepackage[dvipsnames]{xcolor}
\definecolor{cvprblue}{rgb}{0.21,0.49,0.74}
\definecolor{red}{RGB}{192,0,0}
\definecolor{green}{RGB}{0,176,80}
\definecolor{blue}{RGB}{0,112,192}
\newcommand{\cmark}{\ding{51}}
\newcommand{\xmark}{\ding{55}}

\hypersetup{
  colorlinks = true,
  citecolor = green
}
\begin{document}

\definecolor{C1}{HTML}{ECA8A9}
\definecolor{C2}{HTML}{74AED4}
\definecolor{C3}{HTML}{D3E2B7}
\definecolor{C4}{HTML}{CFAFD4}
\definecolor{C5}{HTML}{F7C97E}
\definecolor{C6}{HTML}{E8D5C4}
\definecolor{C7}{HTML}{EEEEEE}
\definecolor{C8}{HTML}{BCEE68}

\definecolor{tabfirst}{rgb}{1, 0.75, 0.7}
\definecolor{tabsecond}{rgb}{1, 0.85, 0.65}
\definecolor{tabthird}{rgb}{1, 0.96, 0.7}

\newcommand{\noo}{\textcolor{red}{\xmark}}
\newcommand{\yes}{\textcolor{OliveGreen}{\checkmark}}

\title{What Is The Best 3D Scene Representation for Robotics? From Geometric to Foundation Models}

% \author{Tianchen Deng\textsuperscript{1}, Yue Pan\textsuperscript{2}, Xingxing Zuo, Shenghai Yuan\textsuperscript{3}, Mingrui Li, Hesheng Wang\textsuperscript{1}, Jingchuan Wang\textsuperscript{1},\\
% Cyrill Stachniss, Lihua Xie\textsuperscript{3}, Danwei Wang\textsuperscript{3}, Weidong Chen\textsuperscript{1} \\
%         % <-this % stops a space
% % <-this % stops a space
% {\textsuperscript{\rm 1} Shanghai Jiao Tong University}
% \textsuperscript{\rm 2} University of Bonn}
% {\textsuperscript{\rm 3} Nanyang Technological University}
% }
\author{Tianchen Deng, Yue Pan,  Shenghai Yuan, Dong Li, Chen Wang, Mingrui Li, Long Chen, 
Lihua Xie, \\  Danwei Wang, Jingchuan Wang, Javier Civera, Hesheng Wang, Weidong Chen
        % <-this % stops a space
% <-this % stops a space
}

\twocolumn[{%
\renewcommand\twocolumn[1][]{#1}%

\maketitle

\begin{center}
  \centering
  \vspace{-15pt}
  \captionsetup{type=figure}
   \begin{overpic}[width=\linewidth]{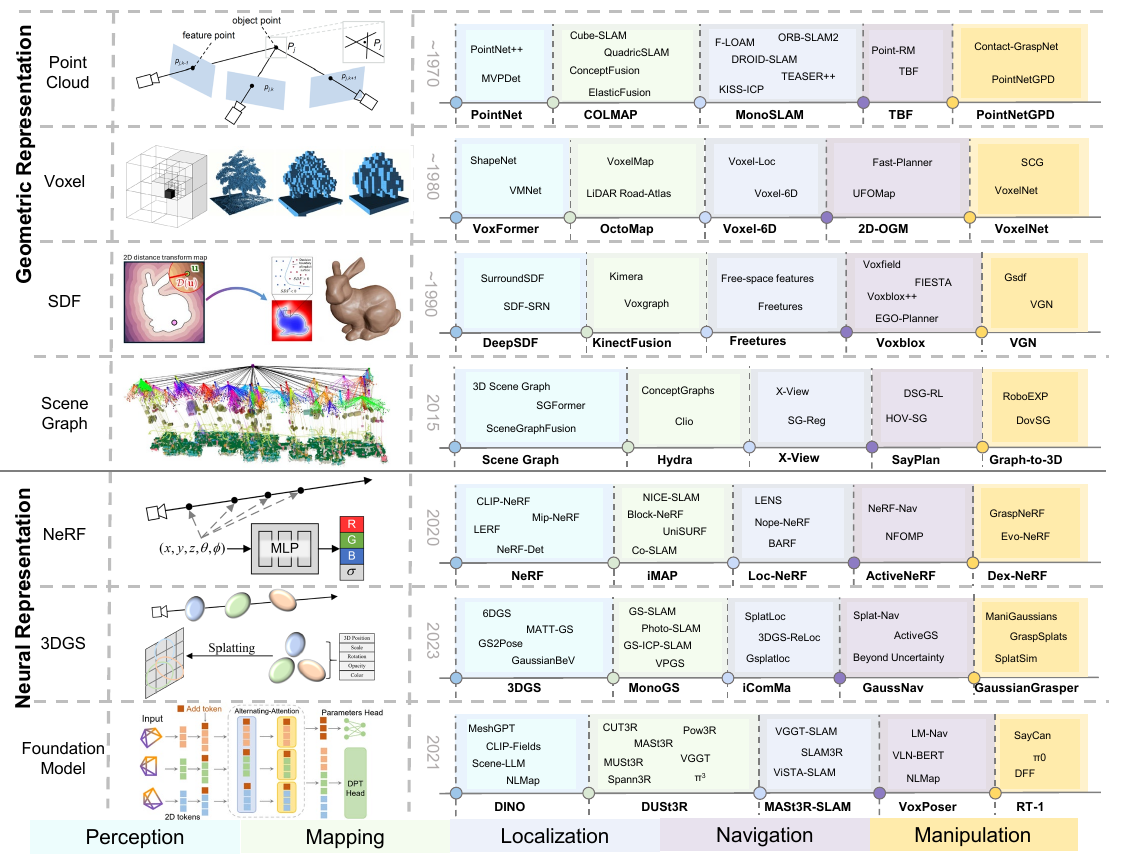}
  \setlength{\abovecaptionskip}{-1pt}
  \vspace{-0.4cm}

      % line 1  
    \put (45.8,71.45) {\tiny{\textbf{\cite{pointnet++}}}} % pointnet++
    \put (45.55,68.95) {\tiny{\textbf{\cite{yin2021mvpdet}}}} % MVPDet
    \put (45.7,65.65) {\scriptsize{\textbf{\cite{pointnet}}}} % pointnet
    \put (55.1,72.8) {\tiny{\textbf{\cite{yang2019cubeslam}}}} % Cube-SLAM
    \put (58.8,71.3) {\tiny{\textbf{\cite{nicholson2018quadricslam}}}} % QuadricSLAM
    \put (56.2,69.74) {\tiny{\textbf{\cite{jatavallabhula2023conceptfusion}}}} % ConceptFusion
    \put (57.1,67.7) {\tiny{\textbf{\cite{elasticfusion}}}} % ElasticFusion
    \put (55.8,65.65) {\scriptsize{\textbf{\cite{schonberger2016structure}}}} % COLMAP
    \put (66.5,72.2) {\tiny{\textbf{\cite{floam}}}} % F-LOAM
    \put (73.9,72.62) {\tiny{\textbf{\cite{orbslam2}}}} % ORB-SLAM2
    \put (70.3,70.6) {\tiny{\textbf{\cite{droidslam}}}} % DROID-SLAM
    \put (73.7,69.1) {\tiny{\textbf{\cite{teaser}}}} % TEASER++ 
    \put (67.25,68) {\tiny{\textbf{\cite{kissicp}}}} % KISS-ICP
    \put (70.95,65.65) {\scriptsize{\textbf{\cite{davison2007monoslam}}}} % MonoSLAM
    \put (80.7,71.5) {\tiny{\textbf{\cite{pointnav2}}}} % Point-RM
    \put (81.1,69.55) {\tiny{\textbf{\cite{pointnav}}}} % TBF
    \put (80.4,65.65) {\scriptsize{\textbf{\cite{pointnav}}}} % TBF 
    \put (93.5,71.8) {\tiny{\textbf{\cite{sundermeyer2021contactgraspnet}}}} % Contact-GraspNet
    \put (93.2,68.85) {\tiny{\textbf{\cite{qi2019pointnetgpd}}}} % PointNetGPD 
    \put (93,65.65) {\scriptsize{\textbf{\cite{qi2019pointnetgpd}}}} % PointNetGPD 

    % line 2
    \put (45.3,61.7) {\tiny{\textbf{\cite{shapenet}}}} % ShapeNet
    \put (47.4,59) {\tiny{\textbf{\cite{hu2022vmnet}}}} % VMNet
    \put (46.95,55.5) {\scriptsize{\textbf{\cite{li2023voxformer}}}} % VoxFormer
    \put (57.5,61.45) {\tiny{\textbf{\cite{yuan2022efficient}}}} % VoxelMap
    \put (59,58.65) {\tiny{\textbf{\cite{wu2023lidar}}}} % LiDAR Road-Atlas
    \put (57.4,55.5) {\scriptsize{\textbf{\cite{octomap}}}} % OctoMap
    \put (68.3,61.5) {\tiny{\textbf{\cite{voxelloc}}}} % Voxel-Loc
    \put (70.3,58.65) {\tiny{\textbf{\cite{voxelloc2}}}} % Voxel-6D
    \put (68.5,55.5) {\scriptsize{\textbf{\cite{voxelloc2}}}} % Voxel-6D
    \put (82.3,61.45) {\tiny{\textbf{\cite{fastplanner}}}} % Fast-Planner
    \put (79,58.65) {\tiny{\textbf{\cite{duberg2020ufomap}}}} % UFOMap
    \put (80.1,55.5) {\scriptsize{\textbf{\cite{elfes2003sonar}}}} % 2D-OGM
    \put (92.2,61.5) {\tiny{\textbf{\cite{varley2017shape}}}} % SCG
    \put (91.6,59) {\tiny{\textbf{\cite{zhou2018voxelnet}}}} % VoxelNet
    \put (92.7,55.5) {\scriptsize{\textbf{\cite{zhou2018voxelnet}}}} % VoxelNet

    % line 3
    \put (47.7,51.2) {\tiny{\textbf{\cite{jiang2023surroundsdf}}}} % SurroundSDF
    \put (48.3,48.7) {\tiny{\textbf{\cite{lin2020sdf}}}} % SDF-SRN
    \put (47.2,45.35) {\scriptsize{\textbf{\cite{deepsdf}}}} % DeepSDF

    \put (56.5,51.2) {\tiny{\textbf{\cite{rosinol2020kimera}}}} % Kimera
    \put (58.8,48.8) {\tiny{\textbf{\cite{reijgwart2019voxgraph}}}} % Voxgraph
    \put (59,45.35) {\scriptsize{\textbf{\cite{kinectfusion}}}} % KinectFusion

    \put (71.8,51) {\tiny{\textbf{\cite{millane2019free}}}} % Free-space features
    \put (70.8,48.65) {\tiny{\textbf{\cite{millane2021freetures}}}} % Freetures
    \put (69.3,45.4) {\scriptsize{\textbf{\cite{millane2021freetures}}}} % Freetures
    
    \put (79.6,52.4) {\tiny{\textbf{\cite{pan2022voxfield}}}} % Voxfield
    \put (84.2,50.8) {\tiny{\textbf{\cite{fiesta}}}} % FIESTA
    \put (81.2,49.45) {\tiny{\textbf{\cite{grinvald2019volumetric}}}} % Voxblox++
    \put (83.1,47.5) {\tiny{\textbf{\cite{zhou2020ego}}}} % EGO-Planner
    \put (82,45.35) {\scriptsize{\textbf{\cite{voxblox}}}} % Voxblox

    \put (90.8,51.3) {\tiny{\textbf{\cite{gsdf}}}} % Gsdf
    \put (93.1,48.9) {\tiny{\textbf{\cite{breyer2021volumetric}}}} % VGN
    \put (91.5,45.35) {\scriptsize{\textbf{\cite{breyer2021volumetric}}}} % VGN

    % line 4
    \put (48.35,41.5) {\tiny{\textbf{\cite{armeni20193d}}}} % 3D Scene Graph
    \put (51.5,39.8) {\tiny{\textbf{\cite{lv2024sgformer}}}} % SGFormer
    \put (50.5,37.8) {\tiny{\textbf{\cite{wu2021scenegraphfusion}}}} % SceneGraphFusion
    \put (49,34.9) {\scriptsize{\textbf{\cite{johnson2015image}}}} % Scene Graph

    \put (63,41.2) {\tiny{\textbf{\cite{gu2024conceptgraphs}}}} % ConceptGraphs
    \put (61.1,38.4) {\tiny{\textbf{\cite{Maggio2024Clio}}}} % Clio
    \put (61,34.9) {\scriptsize{\textbf{\cite{hughes2022hydra}}}} % Hydra

    \put (71.4,41.15) {\tiny{\textbf{\cite{gawel2018x}}}} % X-View
    \put (72.8,38.45) {\tiny{\textbf{\cite{sgreg}}}} % SG-Reg
    \put (72.3,35) {\scriptsize{\textbf{\cite{gawel2018x}}}} % X-View

    \put (83.4,41) {\tiny{\textbf{\cite{ravichandran2022hierarchical}}}} % DSG-RL
    \put (82.1,38.7) {\tiny{\textbf{\cite{werby2024hierarchical}}}} % HOV-SG
    \put (82.7,34.9) {\scriptsize{\textbf{\cite{rana2023sayplan}}}} % SayPlan

    \put (92.7,40.6) {\tiny{\textbf{\cite{roboexp}}}} % RoboEXP
    \put (92.9,38.6) {\tiny{\textbf{\cite{dovsg}}}} % DovSG
    \put (93.5,34.9) {\scriptsize{\textbf{\cite{dhamo2021graph}}}} % Graph-to-3D
    
    % line 5
    \put (46.5,31.4) {\tiny{\textbf{\cite{wang2022clipnerf}}}} % CLIP-NeRF
    \put (50.8,30) {\tiny{\textbf{\cite{mipnerf}}}} % Mip-NeRF
    \put (43.8,28.85) {\tiny{\textbf{\cite{kerr2023lerf}}}} % LERF
    \put (47.7,27.1) {\tiny{\textbf{\cite{xu2023nerfdet}}}} % NeRF-Det
    \put (47.5,24.7) {\scriptsize{\textbf{\cite{NeRF}}}} % NeRF

    \put (61.8,31.7) {\tiny{\textbf{\cite{niceslam}}}} % NICE-SLAM
    \put (60.3,30.25) {\tiny{\textbf{\cite{tancik2022blocknerf}}}} % Block-NeRF
    \put (62.3,28.75) {\tiny{\textbf{\cite{UNISURF}}}} % UniSURF
    \put (59.6,27.05) {\tiny{\textbf{\cite{coslam}}}} % Co-SLAM
    \put (59.6,24.7) {\scriptsize{\textbf{\cite{imap}}}} % iMAP

    \put (69.1,31.4) {\tiny{\textbf{\cite{lens}}}} % LENS
    \put (71.5,29.5) {\tiny{\textbf{\cite{nopenerf}}}} % Nope-NeRF
    \put (70.4,27.55) {\tiny{\textbf{\cite{barf}}}} % BARF
    \put (71,24.7) {\scriptsize{\textbf{\cite{locnerf}}}} % Loc-NeRF

    \put (81,30.8) {\tiny{\textbf{\cite{nerfnavigation}}}} % NeRF-Nav
    \put (81.6,28.2) {\tiny{\textbf{\cite{nfomp}}}} % NFOMP
    \put (82.5,24.7) {\scriptsize{\textbf{\cite{pan2022activenerf}}}} % ActiveNeR

    \put (92.3,30.2) {\tiny{\textbf{\cite{Dai2023GraspNeRF}}}} % GraspNeRF
    \put (92.75,28.3) {\tiny{\textbf{\cite{kerr2022evo}}}} % Evo-NeRF
    \put (92.5,24.7) {\scriptsize{\textbf{\cite{ichnowski2021dexnerf}}}} % Dex-NeRF
    
    % line 6
    \put (44.9,21.3) {\tiny{\textbf{\cite{6dgs}}}} % 6DGS
    \put (50.4,20.05) {\tiny{\textbf{\cite{mattgs2025arxiv}}}} % MATT-GS
    \put (45.8,18.7) {\tiny{\textbf{\cite{gs2pose2024arxiv}}}} % GS2Pose
    \put (50.5,17.2) {\tiny{\textbf{\cite{chabot2024gaussianbev}}}} % GaussianBeV
    \put (47.5,14.6) {\scriptsize{\textbf{\cite{3dgs}}}} %  3DGS

    \put (59.6,21.55) {\tiny{\textbf{\cite{yan2024cvpr}}}} % GS-SLAM
    \put (61.8,20.05) {\tiny{\textbf{\cite{huang2024cvpr-photoslam}}}} % Photo-SLAM
    \put (61.1,18.5) {\tiny{\textbf{\cite{ha2024eccv}}}} % GS-ICP-SLAM
    \put (60.4,16.9) {\tiny{\textbf{\cite{vpgs}}}} % vpgs-SLAM
    \put (59.8,14.6) {\scriptsize{\textbf{\cite{matsuki2024cvpr-monogs}}}} % 

    \put (69.3,21.1) {\tiny{\textbf{\cite{zhai2024splatloc}}}} % SplatLoc
    \put (72.1,19.15) {\tiny{\textbf{\cite{3dgsreloc}}}} % 3DGS-ReLoc
    \put (69.9,17.3) {\tiny{\textbf{\cite{gsplatloc}}}} % Gsplatloc
    \put (69.8,14.6) {\scriptsize{\textbf{\cite{icomma}}}} % iComMa

    \put (79.4,21.2) {\tiny{\textbf{\cite{splatnav}}}} % Splat-Nav
    \put (82.8,19.5) {\tiny{\textbf{\cite{jin2025ral}}}} % ActiveGS
    \put (83.3,17.5) {\tiny{\textbf{\cite{beyondnav}}}} % Beyond Uncertainty
    \put (81.6,14.6) {\scriptsize{\textbf{\cite{gaussnav}}}} % GaussNav

    \put (93.3,21.1) {\tiny{\textbf{\cite{lu2024manigaussian}}}} % ManiGaussians
    \put (94,19.3) {\tiny{\textbf{\cite{ji2024graspsplats}}}} % GraspSplats
    \put (91.4,17.35) {\tiny{\textbf{\cite{qureshi2024splatsim}}}} % SplatSim
    \put (95,14.3) {\scriptsize{\textbf{\cite{zheng2024gaussiangrasper}}}} % GaussianGrasper

    % line 7  
    \put (45.3,11.25) {\tiny{\textbf{\cite{meshgpt}}}} % MeshGPT
    \put (47.5,9.6) {\tiny{\textbf{\cite{clipfields2023}}}} % CLIP-Fields
    \put (46,8.1) {\tiny{\textbf{\cite{li2023scenellm}}}} % Scene-LLM
    \put (47.25,6.55) {\tiny{\textbf{\cite{nlmap}}}} % NLMap
    \put (46,4.1) {\scriptsize{\textbf{\cite{liu2023groundingdino}}}} % DINO

    \put (56,11.25) {\tiny{\textbf{\cite{wang2025cvpr_cut3r}}}} % CUT3R
    \put (63.1,11) {\tiny{\textbf{\cite{jang2025cvpr}}}} % Pow3R
    \put (59.3,9.9) {\tiny{\textbf{\cite{leroy2024eccv}}}} % MASt3R
    \put (56.8,8.2) {\tiny{\textbf{\cite{cabon2025cvpr}}}} % MUSt3R
    \put (62.7,8.5) {\tiny{\textbf{\cite{wang2025cvpr-vggt}}}} % VGGT
    \put (57.3,6.6) {\tiny{\textbf{\cite{wang2025threedv}}}} % Spann3R
    \put (62,6.75) {\tiny{\textbf{\cite{wang2025arxiv-pi3}}}} % PI-3
    \put (60.7,4.1) {\scriptsize{\textbf{\cite{wang2024cvpr-dust3r}}}} % DUSt3R

    \put (74,10.9) {\tiny{\textbf{\cite{maggio2025neurips}}}} % VGGT-SLAM
    \put (74.5,9) {\tiny{\textbf{\cite{liu2025cvpr}}}} % SLAM3R
    \put (74,7.2) {\tiny{\textbf{\cite{zhang2025vista}}}} % ViSTA-SLAM
    \put (74.5,4.1) {\scriptsize{\textbf{\cite{murai2025cvpr-mast3rslam}}}} % MASt3R-SLAM

    \put (83.7,10.7) {\tiny{\textbf{\cite{lm-nav}}}} % LM-Nav
    \put (83.5,8.8) {\tiny{\textbf{\cite{vln-bert}}}} % VLN-BERT
    \put (83.1,6.65) {\tiny{\textbf{\cite{nlmap}}}} % NLMap
    \put (84.2,4.1) {\scriptsize{\textbf{\cite{voxposer}}}} % VoxPoser

    \put (92.8,10.45) {\tiny{\textbf{\cite{saycan}}}} % SayCan
    \put (92.6,8.5) {\tiny{\textbf{\cite{black2024pi_0}}}} % PI-0
    \put (91.6,7.2) {\tiny{\textbf{\cite{shen2023distilled}}}} % DFF
    \put (92.3,4.1) {\scriptsize{\textbf{\cite{brohan2023rt}}}} % RT-1
  \end{overpic}
  
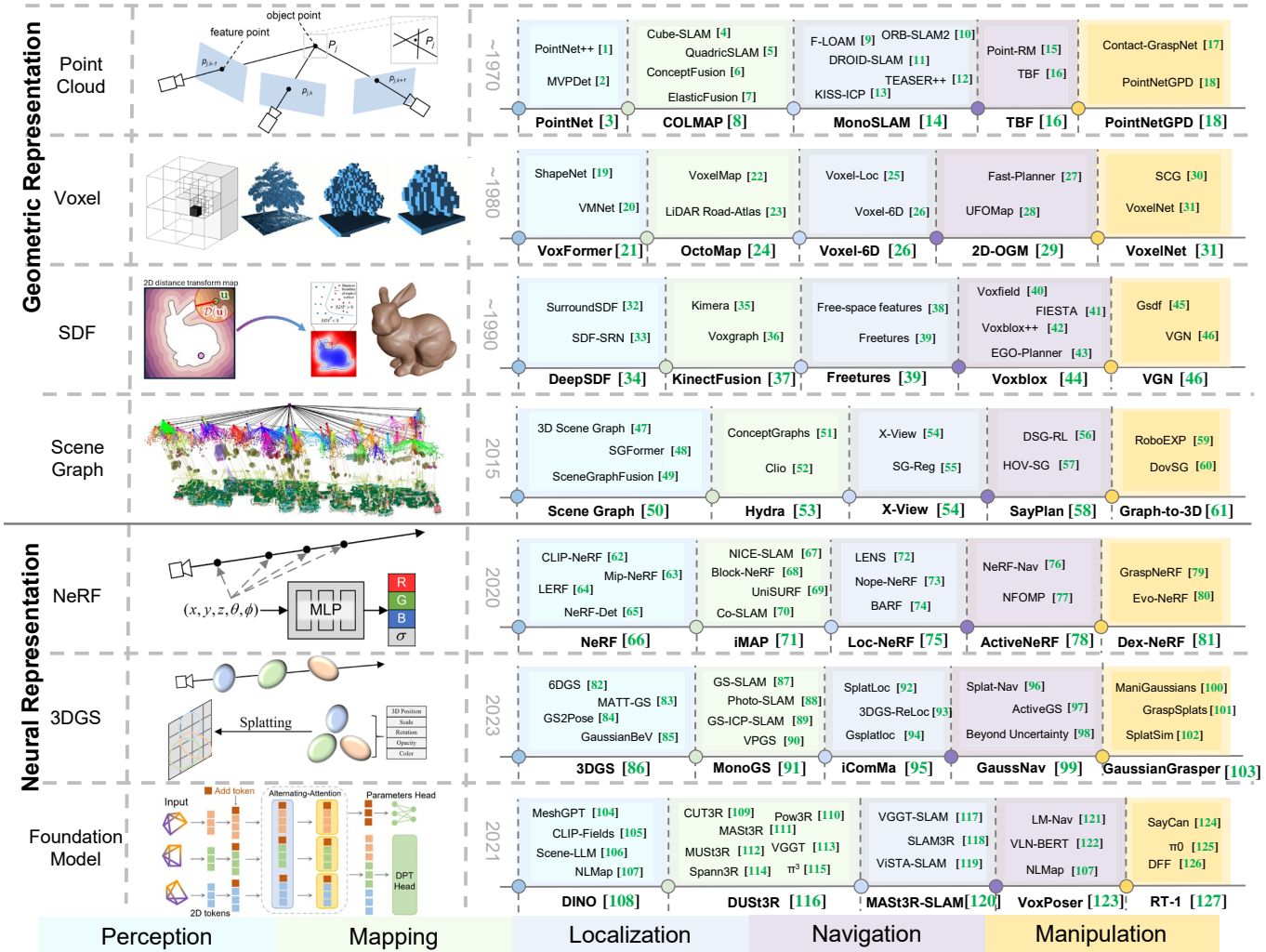
\captionof{figure}{We summarize the development timeline of 3D scene representations in robotics, including point clouds, voxels, meshes, surfels, scene graphs, signed distance fields (SDF), and the most recent representations such as Neural Radiance Fields (NeRF) and 3D Gaussian Splatting (3DGS). Furthermore, we categorize these representations based on their applications across different robotic modules, including mapping, SLAM, localization, planning, manipulation, and simulation. This categorization provides a comprehensive perspective on how different scene representations serve the diverse functional requirements of modern robotic systems. The images of the subfigures are sourced from~\cite{octomap, lin2020sdf, deepsdf, chang2023hydra, wang2025cvpr-vggt}.
  }
  \label{fig:timeline}
  \vspace{-0.2cm}
\end{center}%
}]

\makeatletter{\renewcommand*{\@makefnmark}{}
\footnotetext{Tianchen Deng, Hesheng Wang, Jingchuan Wang, Weidong Chen are with Institute of Medical Robotics and Department of Automation, Shanghai Jiao Tong University, and Key Laboratory of System Control and Information Processing, Ministry of Education, Shang hai 200240, China. Yue Pan is with University of Bonn, Germany. Dong Li and Long Chen are with Institute of Automation, Chinese Academy of Sciences. Chen Wang is with University at Buffalo, Buffalo, NY 14260, USA. Javier Civera is with I3A, University of Zaragoza, Spain. Shenghai Yuan, Lihua Xie, Danwei Wang are with School of Electrical and Electronic Engineering, Nanyang Technological University, Singapore.

% This work is supported by the National Key R\&D Program of China (Grant 2020YFC2007500), the National Natural Science Foundation of China (Grant U1813206), and the Science and Technology Commission of Shanghai Municipality (Grant 20DZ2220400).
}
% The paper headers

\markboth{IEEE Transactions on Robotics}%
{Shell \MakeLowercase{\textit{et al.}}: A Sample Article Using IEEEtran.cls for IEEE Journals}

\IEEEpubidadjcol

% Remember, if you use this you must call \IEEEpubidadjcol in the second
% column for its text to clear the IEEEpubid mark.

\begin{abstract}

% How can robots learn as efficiently as humans and animals? How can robots learn representations of their surroundings through sensor data in a timely manner, similar to humans and animals? 

In this paper, we provide a comprehensive overview of existing scene representation methods for robotics, covering traditional representations such as point clouds, voxels, signed distance functions (SDF), and scene graphs, as well as more recent neural representations like Neural Radiance Fields (NeRF), 3D Gaussian Splatting (3DGS), and the emerging Foundation Models.
While current SLAM and localization systems predominantly rely on sparse representations like point clouds and voxels, dense scene representations are expected to play a critical role in downstream tasks such as navigation and obstacle avoidance. Moreover, neural representations such as NeRF, 3DGS and foundation model are well-suited for integrating high-level semantic features and language-based priors, enabling more comprehensive 3D scene understanding and embodied intelligence.
% This capability is particularly important for embodied intelligence tasks (e.g., cooking or cleaning) and human-robot interaction, where semantic and contextual understanding of the environment is essential for decomposing and executing complex tasks autonomously.
In this paper, we categorized the core modules of the robotics into five parts (Perception, Mapping, Localization, Navigation, Manipulation). We start by presenting the standard formulation of different scene representation methods and compare the advantages and disadvantages of scene representation across different modules. This survey is centered around the question: What is the best 3D scene representation for robotics? We then discuss the future development trends of 3D scene representations, with a particular focus on how the 3D Foundation Model could replace current methods as the unified solution for future robotic applications. The remaining challenges in fully realizing this model are also explored. We aim to offer a valuable resource for both newcomers and experienced researchers to explore the future of 3D scene representations and their application in robotics. 
 We have published an open-source project on GitHub and will continue to add new works and technologies into this project. \href{Github}{https://github.com/dtc111111/awesome-representation-for-robotics}

% 3D Gaussian Splatting (3DGS) has emerged as a significant advancement in the field of computer graphics and 3D vision, achieving high-accuracy novel view synthesis and scene representation without the reliance on neural networks. This innovative approach, characterized by the utilization of numerous 3D Gaussian ellipsoids, represents a significant departure from mainstream neural radiance fields methods. This method has found
% diverse applications in various areas such as virtual reality/augmented reality, metaverse, and robotics. In the context of robotics, where perception and understanding of the environment are pivotal, 3D Gaussian Splatting holds immense promise for improving performance. Given the growing popularity of 3D-GS, this paper presents a comprehensive survey of the state-of-the-art techniques for utilizing 3D Gaussian Splatting to enhance the capabilities of robotics.  
 
\end{abstract}

\begin{IEEEkeywords}
Robotics, 3D scene representation, perception, Localization, navigation, manipulation, Embodied AI, Scene Generation. 
\end{IEEEkeywords}

\section{Introduction}
3D scene representation, which enables a robot to understand its surrounding environment, is fundamental to achieving autonomy and intelligence in robotic systems. It relies on constructing a spatial model of the environment from onboard sensor observations and serves as the foundation for downstream tasks such as navigation, obstacle avoidance, manipulation, and intelligent interaction.

Over the years, significant progress has been made in robotic scene perception and representation, driven by advancements in sensor technologies and algorithmic development. Early robotic state estimation methods relied on processing sensor observations using Kalman filters and expectation-maximization algorithms, and typically represented the environment using 2D grid maps~\cite{gridmap1,gridmap2}. With the advent of 3D sensors such as LiDAR and RGB-D cameras, scene representations have gradually evolved from 2D maps to 3D maps. These 3D representations are commonly constructed using point clouds~\cite{orbslam}, voxels~\cite{voxel}, or meshes~\cite{mesh}.
  
  These approaches are limited to generating discrete scene representations and lack the capacity to produce dense and continuous 3D representation. As a result, they struggle to support complex embodied intelligence tasks, such as robotic navigation and interaction in complex environments. In recent years, deep learning, and computer graphics with robotics has driven substantial progress. Among the myriad techniques
propelling this progress, Neural Radiance Fields (NeRF), 3D Gaussian Splatting and Foundation Model (FM) stand out as particularly promising innovations. Neural Radiance Fields (NeRF) is introduced by Mildenhall et al.~\cite{NeRF}. The core principle of NeRF involves modeling the 3D scene with multi-layer perceptrons (MLP) to map 5D inputs, representing the position and view directions into the scene geometry and appearance. NeRF is becoming increasingly pivotal, significantly enhancing robotic perception and enabling more sophisticated and nuanced interactions between robots and their environments. Several survey papers~\cite{nerfrobotics1,nerfrobotics2,nerfslamsurvey1, neuralsurvey,nerfsurvey} have been published highlighting the advancements in the application of NeRF for robotic systems.
\begin{figure}[t]
    \centering
    \includegraphics[width=\linewidth]{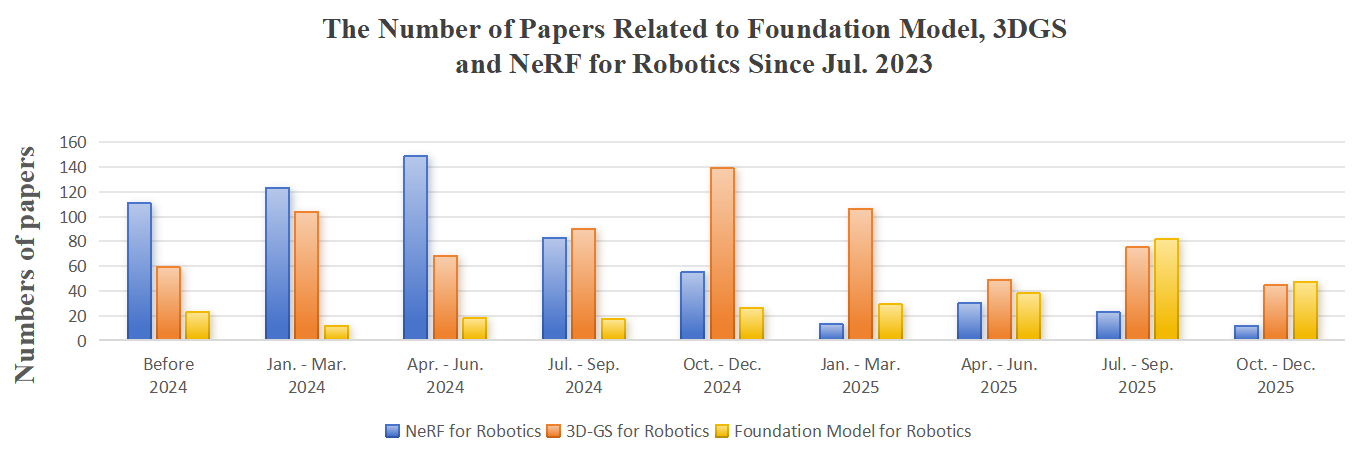}
    \caption{We analyze the trend in the number of papers within the community related to neural scene representation in robotics from Web of Science, including NeRF, 3DGS, and foundation models. We can observe that, over time, the focus has gradually shifted from NeRF to 3DGS and ultimately towards foundation models.
}
    \label{fig:statistics}
    \vspace{-0.4cm}
\end{figure}

\begin{figure*}[t]
    \centering
    \includegraphics[width=\linewidth]{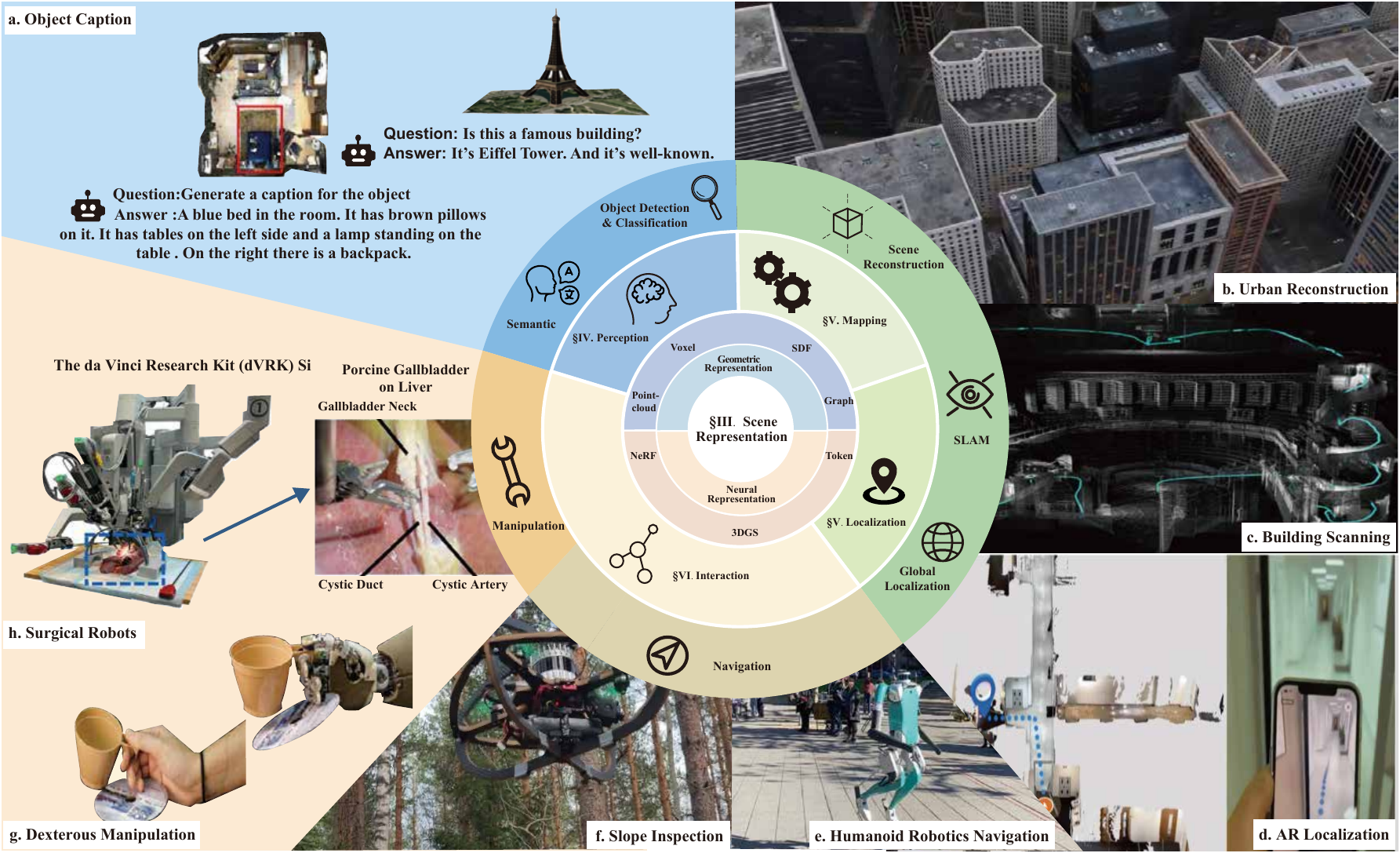}
    \caption{The structure and application of the 3D scene representation for Robotics: 3D Scene Representation ($\S$\ref{sec:general}), Perception ($\S$\ref{sec:perception}), Mapping and Localization module ($\S$\ref{sec:mapping}), and Interaction module ($\S$\ref{sec:interaction}). Subplot(a-h) are extracted from \cite{grounded_3d_llm_2024,octreegs,fast-livo,humanoid,srt}, 
}
    \label{fig:structure}
    \vspace{-15pt}
\end{figure*}
% However, this implementation comes with certain challenges: (i) Computational efficiency. NeRF-Based methods are highly computationally demanding, often requiring significant training time and substantial resources for rendering, particularly for high-resolution outputs. (ii) Intuitiveness, editability, and expressiveness.
% Modifying scenes represented in an implicit representation poses difficulties, as direct alterations to the neural network's weights do not intuitively correspond to changes in the scene's geometric or appearance properties.

In order to further improve the computational efficiency, 3D Gaussian Splatting (3DGS)\cite{3dgs} is designed to
decisively address these bottlenecks, enabling high-quality
real-time (30 fps) scene rendering at 1080p resolution. 3DGS employs an explicit and expressive scene representation that models a scene using millions of learnable 3D Gaussian ellipsoids in space. 
% The design of 3DGS enables rapid reconstruction of high-fidelity 3D models from limited data, effectively overcoming the limitations of traditional sensor-dependent methods. This greatly enhances the potential for applying 3DGS technology in real-world robotic scenarios. 
Due to the emergence of 3DGS as a significant advancement across various fields, many scholars have published extensive reviews on this topic~\cite{3dgssurvey1,3dgssurvey2,3dgssurvey3,3dgsrobotics,spatialperception}.

The tokenize representation (Foundation Model) encodes the entire scene with implicit tokens. By combining the Transformer encoder architecture with large language models (LLMs), it exhibits superior generalization capabilities and sometimes demonstrates an emergent ability to find zero-shot solutions to problems not present in the training data. The Foundation Model has the potential to improve robot capabilities across various modules. Some review papers~\cite{foundationmodel,foundationsurvey,foundationsurvey2} focus on integrating foundation models for robotics autonomy.

 In this paper, our goal is to systematically discuss and categorize the various 3D scene representation methods for robotics, with a particular focus on evaluating which representations are better suited for different modules in robotic systems. We compare the development timeline of various scene representation methods in robotics, as illustrated in Fig.~\ref{fig:timeline}. Additionally, we analyze trends in the number of publications in the robotics community since the introduction of neural scene representations, such as Neural Radiance Fields (NeRF), 3D Gaussian Splatting (3DGS), and tokenizer representation (Foundation Model). As shown in Fig.~\ref{fig:statistics}, there is a clear upward trend, with an increasing number of studies exploring the application of NeRF, 3DGS, and Foundation Model in robotics. We summarize the five core modules essential for real-world robotics in Fig.~\ref{fig:structure}:
%%%%这几个模块需要紧密结合一下nerf和3dgs的应用
\begin{itemize}
    \item 
    %%%
    \textbf{\textit{Perception.}} Robots use this module to perceive their surrounding environment, including semantic segmentation and scene understanding. 
    \item \textbf{\textit{Mapping}.} This module is used to model and generate the map of the surrounding environment with the sensors on robots. 
    \item \textbf{\textit{Localization}.} Robot localization is the process of determining where a  robot is located with respect to its environment. 
     \item \textbf{\textit{Manipulation}.} Robot manipulation refers to the ways robots interact with the objects around them: grasping an object, opening a door, packing an order into a box. All these actions require robots to plan and control the motion of their hands and arms in an intelligent way.
    \item \textbf{\textit{Navigation.}} Robot navigation refers to the robot's ability to identify its position within a reference frame and plan a path toward a target location.
   
    % \item \textbf{\textit{Simulation}.} Robotics simulation enables virtual training and programming that can use physics-based digital representations of environments, robots, machines, objects and other assets. The latest simulators can generate datasets to be used to train machine learning models that will run on the physical robots.
\end{itemize}

In Sec.~\ref{sec:background}, we start with the background of different scene representation methods. We provide a brief introduction to the problem formulation of different scene representation, followed by a comparison of their differences. We then discuss improvements to the scene representation in Sec.~\ref{sec:general}: (i) \textbf{Memory-efficiency:}  Acceleration and compression for the scene representation.  
(ii) \textbf{Photorealism}
Rendering and visualization ability of different representations.
(iii) \textbf{Geometric Representation Capability} Geometric representation capability of different scene representations.

Moving on to the robotic perception module, there are two parts in this module (Sec.~\ref{sec:perception}): (i) \textbf{Object Detection}: Object-level perception focuses on modeling and identifying individual objects within a scene, including their categories, attributes, and spatial extents. (ii) \textbf{Scene Understanding}: Beyond isolated objects, scene-level perception aims to capture the holistic structure of the environment by modeling spatial layouts, semantics, relationships among objects, and the overall scene context. 

% (ii) \textbf{Multi-Modal:} By utilizing multimodal sensor inputs, we achieve multimodal perception of indoor scenes and enable effective information fusion. (iii) \textbf{Complex environments:} For modeling complex weather conditions or scenes with varying lighting, 3DGS creates explicit scene reconstructions, effectively decoupling different environmental factors, thereby facilitating the modeling of complex weather scenarios. This greatly enhances the robot's perception capabilities in real-world scenarios.

In the mapping and localization module (Sec.~\ref{sec:mapping}), existing methods have achieved promising results in the areas of SLAM and localization. Neural scene representations enable more accurate and denser modeling of the environment, which is particularly beneficial for obstacle avoidance. This capability is crucial for robots in navigation and manipulation. There are three parts: (i) \textbf{Scene Reconstruction:} The map reconstruction capability of scene representation includes geometric accuracy and rendering quality, with reconstruction abilities in static scenes, large-scale outdoor scenes, and dynamic scenes. (iii) \textbf{SLAM} The SLAM section mainly includes the map accuracy, pose accuracy, and real-time performance of different scene representation methods during the SLAM process. (iv) \textbf{Global Localization}: Global localization mainly involves the accuracy and real-time performance of localization when using an existing map.  

In the manipulation module (Sec.~\ref{sec:manipulation}), we mainly compare the grasping frameworks based on different scene representation methods. Traditional methods offer higher real-time performance and computational efficiency in grasping, but they are limited in generalization and handling complex object manipulation tasks. In contrast, neural-based scene representations offer certain capabilities in generating novel viewpoints and generalizing across multiple scenes, making them more adaptable to complex tasks. Foundation model-based approaches enable zero-shot grasping tasks, providing strong generalization abilities. Furthermore, the integration of language information allows these models to support interactive grasping and enhances their ability to understand and plan higher-level cognitive tasks

 In the navigation module (Sec.~\ref{sec:interaction}), compared to traditional scene representation methods, neural scene representations provide highly accurate reconstructions of the environment. In addition, they facilitate better integration with semantic and language information, enabling more complex navigation tasks. We divide the navigation module into two components: (i) \textbf{Planning}: Generating an optimal or feasible path from the current location to a target destination while avoiding obstacles.   (ii) \textbf{Exploration}: Actively navigating and mapping previously unknown areas.

In this paper, we explore the most suitable 3D scene representation for different modules of robotics, examining the approaches, benchmarking performance, and discussing challenges and future directions. 
Our main contributions are listed below:
% \begin{figure}[t]
%     \centering
%     \includegraphics[width=\linewidth]{img/figure2-4.png}
%     \caption{ NeRF compared with 3DGS. (a) NeRF samples multiple points along each ray and queries the MLP to get the corresponding colors and opacities, which can be interpreted as a form of backward mapping. (b) In contrast, 3DGS use numerous 3D gaussian ellipsoids to represent the scene. It projects all 3D Gaussians into the image space and then performs parallel rendering with splatting and rasterization. The two figures at the bottom are from NeRF~\cite{NeRF} and 3DGS~\cite{3dgs}, respectively.
% }
%     \label{fig:nerf3dgs}
% \end{figure}

\begin{figure}[t]
    \centering
    \includegraphics[width=\linewidth]{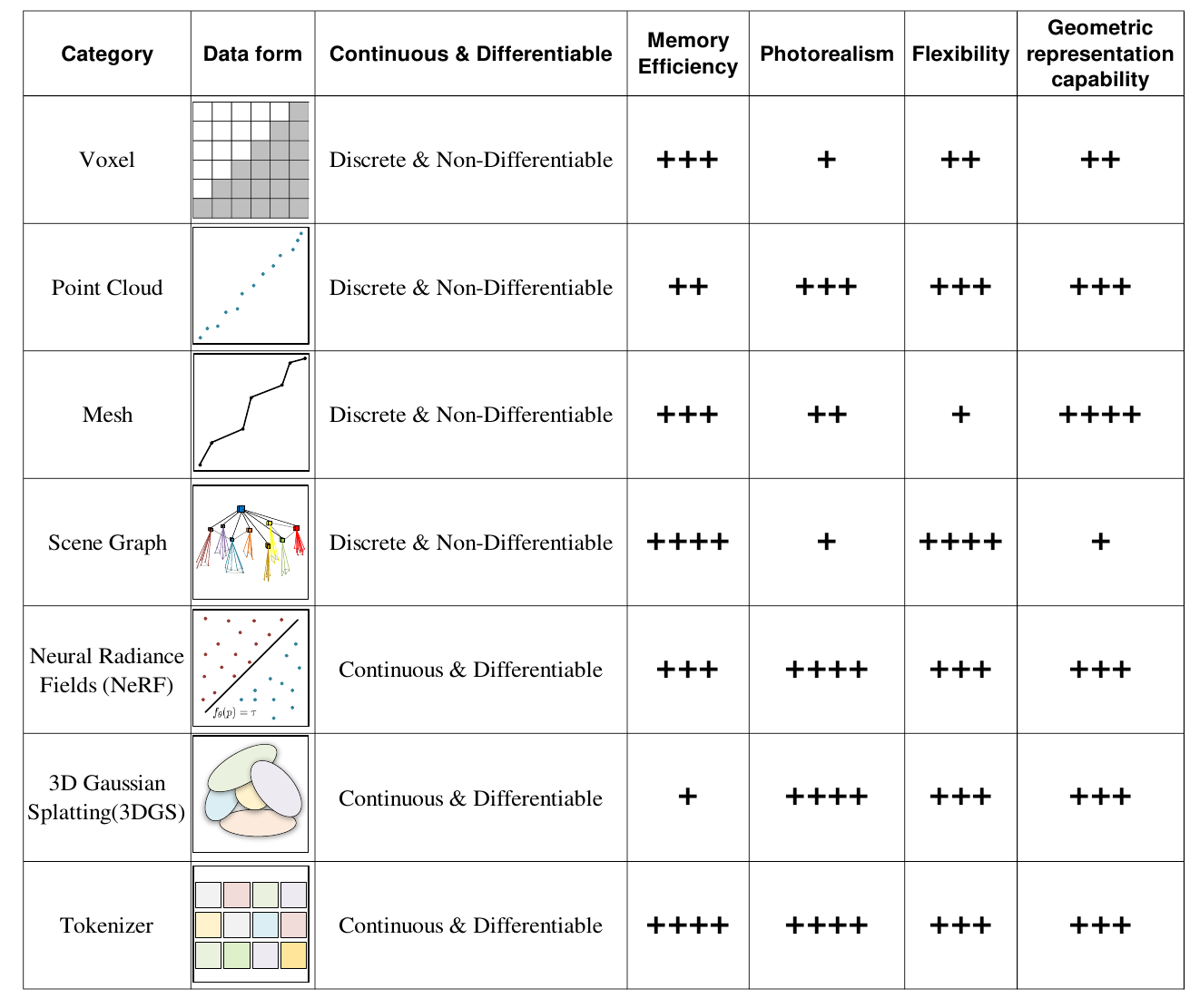}\vspace{-5pt}
    \caption{ We compare various scene representations across several dimensions, including data form, continuity, memory usage, photorealism, flexibility, and geometric representation capability. Exact values differ, but the overall trend is indicative and remains comparable across methods.
}
    \label{fig:scenecompare}\vspace{-20pt}
\end{figure}

\begin{itemize}
    \item \textbf{Comprehensive, up-to-date review and benchmark:} This paper offers an extensive and current review of different scene representation for robotics, encompassing both classical and cutting-edge approaches. For each module, we provide a detailed introduction and highlight the advantages of different scene representation in this module. To the best of our knowledge, this is the first review focused on the discussion in robotics.
\item \textbf{Future directions for 3D scene representation:} In each module of robotics, we identify the technical limitations of current research and suggest several promising directions for future work, aiming to inspire further advancements in this rapidly evolving field.
\item \textbf{Open-source project:} We have published an open-source project on GitHub
that compiles different scene representation in robotics related articles, and will continue to add new works and technologies into this project 
\href{https://github.com/dtc111111/awesome-representation-for-robotics}{https://github.com/dtc111111/awesome-representation-for-robotics}. 
We hope that more researchers can use it to access the latest
research information.
\end{itemize}

\section{Background and Problem Formulation}
\label{sec:background}
In this section, we provide a brief formulation of traditional scene representation and neural scene representation. 

\noindent\textit{\textbf{Point Cloud}}
A point cloud is an unordered set of \( N \) 3D points \( P = \{p_i\} \; | \; p_i \in \mathbb{R}^3\) for \( i = 1, \ldots, N \), representing the surface of an object or scene. Point clouds are sparse, unstructured, and typically derived directly from depth sensors, LiDAR, stereo matching, structure-from-motion and SLAM~\cite{chen1989representation}.

\noindent\textit{\textbf{Voxel Grid}}
A voxel-grid representation models the modalities of interest (e.g., density, color, or feature) explicitly in its grid cells~\cite{frieder1985back}. Such an explicit scene representation is efficient to query for any 3D positions via interpolation:
% ~\cite{sun2022direct}

\begin{equation}
\text{interp}(x, V) : (\mathbb{R}^3, \mathbb{R}^{C \times N_x \times N_y \times N_z}) \to \mathbb{R}^C,
\end{equation}
where $x$ is the queried 3D point, $V$ is the voxel grid, $C$ is the dimension of the modality, and $N_x \cdot N_y \cdot N_z$ is the total number of voxels. Voxel grids enable scene or object representation with adjustable granularity by tuning grid resolution.

\noindent\textit{\textbf{SDF}}
% 在这里提一下 TSDF和mesh
A signed distance function is a continuous function that, for a given spatial point, outputs the point's distance to the closest surface, whose sign encodes whether the point is inside (negative) or outside (positive) of the watertight surface:
\begin{equation}
SDF(x) = s : x \in \mathbb{R}^3, s \in \mathbb{R}.
\end{equation}
The underlying surface is implicitly represented by the iso-surface of \(SDF(\cdot) = 0\). A view of this implicit surface can be rendered through raycasting or rasterization of a mesh obtained with, for example, Marching Cubes~\cite{bloomenthal1997introduction}.
% deepsdf

% johnson2015image
\noindent\textit{\textbf{Scene Graph}}
% 2d and 3D
Formally, the scene graph \( SG \) is a directed graph data structure; this can be defined in the form of a tuple \( SG = (O, R, E) \), where \( O = \{o_1, \ldots, o_n\} \) is the set of objects detected in the images, and \( n \) is the number of objects. Each object \( o_i \) is associated with a set \( A_i \), where \( c_i \) and \( r_i \) represent the category and attributes of the object respectively. \( R \) stands for a set of relationships between objects, where the relationship between the \( i \)-th and the \( j \)-th object can be expressed as \( r_{i \rightarrow j}, i, j \in \{1, \ldots, n\} \). \( E \subset O \times R \times O \) represents the edges between the object instances and the relationship nodes, so there are at most \( n \times n \) edges in the initial graph. Then, \( Edge(o_i, r_{i \rightarrow j}) \) is classified as irrelevant~\cite{johnson2015image}.
Specifically, a 3D Scene Graphs is a hierarchical multigraph \( G = (V, E) \) in which the set of vertices \( V \) comprises \( V_1 \cup V_2 \cup \ldots \cup V_K \), with each \( V_k \) signifying the set of vertices at a particular level of the hierarchy \( k \). Edges stemming from a vertex \( v \in V_k \) may only terminate in \( V_{k-1} \cup V_k \cup V_{k+1} \), i.e., edges connect nodes within the same level, or one level higher or lower~\cite{armeni20193d}.

\noindent\textit{\textbf{Neural Radiance Fields}}
\label{sec:nerf}
NeRF encodes 3D scenes into the weights of MLP. The original NeRF~\cite{NeRF} use a 5D vector $(x,y,z,\theta,\phi)$ as input. It includes 3D coordinate vector $\mathbf{x}=(x,y,z)$ and a 2D direction vector $\mathbf{d}=(\theta,\phi)$. The output of the network is RGB vector $\mathbf{c}=(r,g,b)$ and a volume density $\sigma$. The scene representation of NeRF can be summarized as:
\begin{equation}
    F_{\theta}(\mathbf{x},\mathbf{d})= (\mathbf{c},\sigma)
\end{equation}
where $\theta$ denotes the parameter of the network. The original NeRF applies positional encoding method to improve the quality, which can map the input vector into a high-dimensional space to better represent the high-frequency information:
\begin{equation}
\begin{split}
\gamma(p)=\left(\sin \left(2^0 \pi p\right), \cos \left(2^0 \pi p\right), \ldots, \sin \left(2^{L-1} \pi p\right), \right. \\ \left.  \cos \left(2^{L-1} \pi p\right)\right)
\end{split}
\end{equation}
where L is the hyperparameter. The volume rendering is an integration process. This process involves integrating the color and density of all sampled points along a ray into a single pixel on the target image, along the viewing direction:
\begin{equation}
    C(\mathbf{r})=\int_{t_n}^{t_f} T(t) \sigma(\mathbf{r}(t)) \mathbf{c}(\mathbf{r}(t), \mathbf{d}) d t
\end{equation}
where $t_n$ and $t_f$ are near and far bounds of the camera ray $\mathbf{r}=\mathbf{o}+t\mathbf{d}$. The transmittance is calculated as:
\begin{equation}
    T(t)=\exp \left(-\int_{t_n}^t \sigma(\mathbf{r}(s)) d s\right)
\end{equation}
Then, NeRF discretizes the integration process:
\begin{equation}
\hat{C}(\mathbf{r})=\sum_{i=1}^N T_i \alpha_i \mathbf{c}_i,  T_i=\exp \left(-\sum_{j=1}^{i-1} \sigma_j \delta_j\right)
\end{equation}
where $\alpha_i=\left(1-\exp \left(-\sigma_i \delta_i\right)\right)$, $\delta_j=t_{j+1}-t_j$ is the distance between two adjacent samples. 
%%%对应SDF-nerf
% When using the signed distance function~\cite{sdf,deepsdf}, the equation can formulate as:
% \begin{equation}
% \hat{C}(\mathbf{r})=\frac{1}{\sum_{i=1}^N w_i} \sum_{i=1}^N w_i \mathbf{c}_i, \quad w_i=\sigma\left(\frac{s_i}{t r}\right) \cdot \sigma\left(-\frac{s_i}{t r}\right)
% \end{equation}
% where $\sigma(\cdot)$ is the sigmoid function and $tr$ denotes the truncation distance.

\noindent\textit{\textbf{3D Gaussian Splatting}}
% \begin{figure*}[t]
%     \centering
%     \includegraphics[width=\linewidth]{img/figure5-1.png}
%     \caption{ An illustration of the different aspects of future direction of the neural scene representation, including Photorealistic Rendering, Sparse View Representation, Acceleration \& Compression, and Geometry \& Physics.
% Subplot(a) to subplot(d) are extracted from \cite{mipsplat}, \cite{pixelsplat}, \cite{compactgs1}, \cite{2dgs}, respectively.
% }
%     \label{fig:general}
% \end{figure*}
3DGS~\cite{3dgs} present an explicit radiance field-based scene representation that represent a radiance field using a large number of 3D anisotropic ellipsoids.
 The 3D Gaussian ellipsoids are defined by a full 3D covariance matrix $\sum$ defined in world space centered at point $\mu$:
\begin{equation}
    G(\mathbf{x})=e^{-\frac{1}{2}(\mathbf{x})^T \Sigma^{-1}(\mathbf{x})}
\end{equation}
This Gaussian is multiplied by $\alpha$ in our blending process. The projection formulation from 3D Gaussian into camera coordinates is given as follows:
\begin{equation}
    \Sigma^{\prime}=\mathbf{J} \mathbf{W} \Sigma \mathbf{W^T} \mathbf{J^T}
\end{equation}
where $W$ denotes the viewing transformation, $\sum^{'}$ denotes the covariance matrix. $J$ is the Jacobian of the affine approximation of the projective transformation. To ensure the semi-positive definiteness of the covariance matrix, 3DGS reparameterized the covariance matrix as a combination of a rotation matrix $R$ and a scaling matrix $S$:
\begin{equation}
    \mathbf{\Sigma}=\mathbf{R S} \mathbf{S}^T \mathbf{R}^T
\end{equation}
The reparameterization method constrain the scaling vector to give Gaussian primitives a flattened characteristic. In addition to geometric attributes, each Gaussian primitive also stores an opacity value $\alpha$ and a set of learnable Spherical Harmonic (SH) parameters to represent view-dependent appearance.

\noindent\textit{\textbf{Foundation Model}}
A Tokenizer Scene Representation is a method that encodes complex scene information into a set of discrete tokens, facilitating efficient processing and understanding of spatial data. It transforms visual, geometric, or semantic elements of a scene (e.g., objects, textures, or spatial relationships) into a sequence of tokens, typically modeled using attention mechanisms or neural networks~\cite{bommasani2021opportunities, firoozi2025foundation}. This representation is widely used in computer vision, 3D scene understanding, and multimodal learning, enabling structured scene encoding, reduced computational complexity, and unified cross-modal representations:
\begin{equation}
    TSR(x) = \{t_1, t_2, \dots, t_n\} : x \in \mathbb{R}^k, t_i \in \mathbb{R}^d,
\end{equation}
where \( x \) denotes the input scene data, and \( \{t_i\} \) is the tokenized output. Scenes can be reconstructed or queried by decoding these tokens using learned mappings or generative models.

\section{3D Scene Representation}
\label{sec:general}

%在这个地方补一个表格，说明不同场景表示的一个对比，然后落到Neural场景表示的发展趋势，做一个neural 场景表示的一个survey
In this section, we summarize the development timeline of 3D scene representations, as shown in Fig. \ref{fig:timeline}. This includes traditional scene representation methods such as point clouds, voxels, SDF, and scene graph, as well as neural scene representations. We then compare these different representations across several dimensions, including data form, continuity, memory usage, granularity, flexibility, and geometric representation capability, as illustrated in Fig. \ref{fig:scenecompare}. Finally, we discuss four key points for 3D scene representations: (i) Memory efficiency, (ii) Photorealism, (iii) Geometric Representation Capability. 
%%%这一章需要找到，认为场景表示重要的特点，首先认为photorealism非常重要，其次是data efficiency和泛化性，第三点是memory efficiency 第四点是几何表示能力

\subsection{Memory Efficiency} 
Memory efficiency is paramount for robot's scene representation, particularly when considering the need for lifelong operation and deployment on edge computing platforms with limited resources. This section critically examines the memory footprint of various scene representation methods, highlighting recent advancements aimed at enhancing memory efficiency. Key improvements include optimized data structures, algorithmic compressions, and model pruning techniques, which collectively enable more scalable and efficient scene representations suitable for long-term, resource-constrained robotic applications. 
%%%%%需要找一下各个方法的compression的论文 memory efficient, pointcloud和voxel一起来说，point采样之后就是voxel，提高速度，降低精度，sdf的表示

\noindent\textbf{\textit{Geometric Representations}}
Point cloud scene representation, exhibits relatively low memory efficiency, especially in scenarios where robots operate over extended lifetimes, leading to the accumulation of points in the scene. Several methods have been proposed to address this issue, with a focus on point cloud compression and pruning. ~\cite{pointsurvey} provides a comprehensive survey of point cloud compression techniques. \cite{kdtree} introduces a KD-tree-based approach for point cloud sampling and \cite{binarytree} combinations of binary trees and quadtrees for point compression. \cite{octree1} explore downsampling point clouds through voxel-based or employ Octree methods for point cloud downsampling, which effectively reduces memory usage. \cite{lod1} utilizes an LOD structure for point cloud compression, introducing a general point cloud encoder that constructs an voxel-based LOD structure. \cite{meshcompression} apply mesh-based representation for compressing point clouds. 

\noindent\textbf{\textit{NeRF-Based Representations}}
NeRF represents the scene with a radiance field using a neural network, resulting in relative lower memory usage compared to point clouds, though at the cost of slower processing speed. Plenoxel~\cite{plenoxels} voxelizes the radiance fields and stores a scalar for density along with spherical harmonics coefficients for direction-dependent color. TensoRF~\cite{tensorf} improves memory efficiency by storing scalar density and vector features as factorized tensors, which can be decoded via MLP or used with SH coefficients.

\noindent\textbf{\textit{3DGS-Based Representations}}
3DGS requires millions of distinct Gaussian primitives to represent the geometry and appearance of a scene, leading to significant storage overhead. For example, a small room scene may require millions of Gaussian spheres. Some methods achieve compression and acceleration by directly removing redundant 3D ellipsoids. \cite{compactgs1,compactslam} remove the redundant 3D Gaussian ellipsoids with a novel volume-based masking strategy. 
CompactGS~\cite{compactgs1} introduces a residual vector quantization codebook to compress different attributes into codebooks. HAC~\cite{hac} introduces a hash-grid assisted context framework for a highly compact 3DGS representation. It employs a binary hash grid to enforce continuous spatial consistency and leverages Gaussian distributions to accurately estimate the probabilities of each quantized attribute.
Scaffold-GS~\cite{scaffoldgs} introduces a structure-based 3D Gaussian scene representation that employs anchor points to distribute local 3D Gaussians.

\noindent \textbf{\textit{Foundation Model}}
Token-based scene representations require less memory than point cloud representations but typically demand more network parameters. Token-based scene representations are typically extracted through transformer-based architectures.  Numerous efforts have been made to mitigate the resource cost of attention-based transformer architectures. Some methods, such as Longformer~\cite{longformer} and BIGBIRD~\cite{bigbird}, decompose conventional attention into local windowed attention and task-specific global attention, effectively reducing self-attention complexity to linear. HEPOS~\cite{hepos} introduces head-wise positional strides, allowing each attention head to focus on a specific subset of the input sequence. Other methods, such as knowledge distillation~\cite{kd1}, quantization~\cite{quantization} are also employed to improve memory efficiency.

\subsection{Photorealism}
 High-fidelity and photorealistic reconstructions play a crucial role in robotic teleoperation, perception, and human-robot interaction. Traditional scene representations are generally sparse and cannot represent scenes in a continuous manner. Neural scene representation is anticipated to enable realistic rendering across various scenarios. 
 In contrast to traditional representations, the continuity of neural scene representations enables photorealistic rendering, which is particularly beneficial for tasks such as view synthesis, simulation, and human-robot interaction.
 % The illustration of methods focus on photorealistic are presented in Fig.~\ref{fig:photorealistic}. Comparison results are presented in Tab.~\ref{tab:photo}.

\noindent\textbf{\textit{Geometric Representations}}
Point-Based Computer Graphics~\cite{pointrender} introduced point-based rendering, treating points as primitives and enabling high-fidelity 2D synthesis from raw point clouds. Volume rendering~\cite{volumerender} established the classical voxel-based paradigm, where 3D voxel grids are integrated along viewing rays to generate accurate 2D synthesis. This approach is the foundation of modern volumetric rendering techniques. Building upon point-based methods, Surface Splatting~\cite{surfacesplatting} proposed a technique that renders opaque and transparent surfaces from point clouds without requiring explicit connectivity. The method uses a screen-space formulation of the Elliptical Weighted Average (EWA) filter, which mitigates aliasing and fills gaps in sparse point clouds, enabling smooth, continuous surface rendering.

\noindent\textbf{\textit{NeRF-Based Representations}}
Mip-NeRF~\cite{mipnerf} improves anti-aliasing and geometric consistency by introducing a multi-scale representation. It efficiently renders anti-aliased conical frustums instead of rays, and reduces objectionable aliasing artifacts and significantly improves NeRF's ability to represent fine details.
Mip-NeRF 360~\cite{mipnerf360} further extends NeRF to unbounded, real-world scenes. It uses a non-linear scene parameterization, online distillation, and a novel distortion-based regularizer to overcome the challenges presented by highly intricate and unbounded real-world scenes. 

\noindent\textbf{\textit{3DGS-Based Representations}}
Due to its discrete sampling paradigm, which treats each pixel as a single point rather than an area, 3DGS is prone to aliasing when handling varying resolutions, resulting in blurring or jagged edges. Multi-scale 3DGS~\cite{multi3dgs} proposes a multi-scale 3D Gaussian splatting algorithm, which maintains Gaussians at different scales to represent the same scene. Mip-splatting~\cite{mipsplat} also focuses on the multi-scale rendering challenge. It introduces a Gaussian low-pass filter based on Nyquist's theorem to constrain the frequency of the 3D Gaussians according to the maximal sampling rate across all observed samples. 
% FreGS~\cite{fregs} introduces a progressive frequency regularization method to address the over-reconstruction issue in the frequency domain. It performs coarse-to-fine Gaussian densification by leveraging low-to-high frequency components, which are easily extracted using low-pass and high-pass filters in the Fourier space. RDO-Gaussian~\cite{rdogaussian} proposes an end-to-end rate-distortion optimization framework with dynamic pruning and entropy-constrained vector quantization. It also improves the model the colors of different regions and materials with learnable numbers of parameters. 
% \begin{figure}[t]
%     \centering
%     \includegraphics[width=\linewidth]{img/figure8-1.png}
%     \caption{ An illustration of the different methods to improve the rendering performance. Images are from \cite{mipsplat, gaussianpro, gs-shader}.
% }
%     \label{fig:photorealistic}
% \end{figure}

\noindent\textbf{\textit{Foundation Model}}
Tokenizer-based models aim to directly synthesize novel views through feed-forward inference without relying on explicit 3D representations such as NeRF or 3DGS. We categorize foundation model-based approaches into two types: reconstruction-based and generation-based.
For reconstruction-based methods, SRT~\cite{srt} designs a representative encoder-decoder architecture as the scene representation transformer, where the encoder encodes the input images into tokens, and the transformer-based decoder outputs light field rays for novel view synthesis. OSRT~\cite{osrt} targets object-centric 3D scenes and extends SRT by incorporating a slot attention module that maps the encoded latent representations into object-centric slot representations. For generation-based methods, earlier works mainly adopt transformer-based autoregressive models. For instance, GFVS employs a VQGAN~\cite{vqgan} combined with a transformer to autoregressively model the distribution. More recently, latent diffusion models~\cite{ladiffusion} have been widely applied to novel view synthesis due to their ability to generate high-resolution images. It encodes input images into a latent space using a pretrained variational autoencoder and apply diffusion within this latent representation.

\begin{table*}[]
\caption{List of commonly used datasets related to scene representation for robotics}
\vspace{-5pt}
\scalebox{0.78}{
\setlength{\tabcolsep}{0.5mm}{
\begin{tabular}{lcccccccccccc}
\toprule
\multirow{2}{*}{Dataset} & \multirow{2}{*}{Year} & \multirow{2}{*}{Scenes} & \multicolumn{4}{c}{Senor Modalities} & \multicolumn{3}{c}{GroundTruth}        & \multirow{2}{*}{GT Method} & \multirow{2}{*}{Description}                                                                                                    & \multirow{2}{*}{Corresponding Robot Module}                                         \\ 
                         &                       &                         & RGB  & Depth  & Lidar  & IMU  & Pose & 3D Map & \multicolumn{1}{c}{Label} &                            &                                                                                                                                 &                                                                                       \\ \hline
\rowcolor{C7!50}Tanks and Temples~\cite{tanks}        & ToG'17                & Indoor/Outdoor          &  \color{ForestGreen}{\cmark}           &   \color{BrickRed}{\xmark}     &  \color{BrickRed}{\xmark}      &  \color{BrickRed}{\xmark}    &  \color{ForestGreen}{\cmark}    &  \color{ForestGreen}{\cmark}   &    \color{BrickRed}{\xmark}                       & Laser Scan,ICP             & \begin{tabular}[c]{@{}c@{}}High-quality indoor and outdoor Mesh by \\ industrial laser scanner\end{tabular}                     & Representation, Mapping                                                               \\
\rowcolor{C7!50}DeepBlending~\cite{deepblending}             & TOG'18                & Indoor/Outdoor          & \color{ForestGreen}{\cmark}            &  \color{BrickRed}{\xmark}      &  \color{BrickRed}{\xmark}      &  \color{BrickRed}{\xmark}    &  \color{ForestGreen}{\cmark}     & \color{BrickRed}{\xmark}     &   \color{BrickRed}{\xmark}                         & COLMAP                     & \begin{tabular}[c]{@{}c@{}}Images with sufficient diversity \\ in scene content\end{tabular}                                    & Representation                                                                        \\
\rowcolor{C7!50}RealEstate10k~\cite{RealEstate10k}            & ToG'18                & Indoor/Outdoor          &  \color{ForestGreen}{\cmark}           &  \color{BrickRed}{\xmark}      &  \color{BrickRed}{\xmark}      &  \color{BrickRed}{\xmark}    &  \color{ForestGreen}{\cmark}    &  \color{BrickRed}{\xmark}   &  \color{BrickRed}{\xmark}                         & COLMAP                     & \begin{tabular}[c]{@{}c@{}}Video clips on YouTube captured \\ from a moving camera\end{tabular}                                 & Representation  \\                                            
\rowcolor{C7!50} DL3DV-10K~\cite{dl3dv}             & CVPR'24               & Indoor/Outdoor          &  \color{ForestGreen}{\cmark}           &  \color{BrickRed}{\xmark}      &  \color{BrickRed}{\xmark}      &  \color{BrickRed}{\xmark}    & \color{ForestGreen}{\cmark}     & \color{BrickRed}{\xmark}    &      \color{BrickRed}{\xmark}                     & COLMAP                     & \begin{tabular}[c]{@{}c@{}} A large-scale scene dataset covering \\ both bounded 
and unbounded scenes,\\ with different levels of reflection,
transparency.\end{tabular}                         & Representation, Mapping \\ \hline

\rowcolor{C4!20}KITTI~\cite{kitti}                    & CVPR'12               & Indoor/Outdoor          &   \color{ForestGreen}{\cmark}         &  \color{BrickRed}{\xmark}      & \color{ForestGreen}{\cmark}       & \color{ForestGreen}{\cmark}     & \color{ForestGreen}{\cmark}     &   \color{BrickRed}{\xmark}  &  \color{ForestGreen}{\cmark}                         & GNSS/INS                   & \begin{tabular}[c]{@{}c@{}}A city-scale dataset created for \\ autonomous driving research\end{tabular}                         & \begin{tabular}[c]{@{}c@{}}Perception, Mapping, \\ Localization\end{tabular}                                                         \\
\rowcolor{C4!20}TUM RGB-D~\cite{tum}               & IROS'12               & Indoor                  & \color{ForestGreen}{\cmark}            &  \color{ForestGreen}{\cmark}      &  \color{BrickRed}{\xmark}      &  \color{BrickRed}{\xmark}     & \color{ForestGreen}{\cmark}     &  \color{BrickRed}{\xmark}    &  \color{BrickRed}{\xmark}                          & Motion Capture             & \begin{tabular}[c]{@{}c@{}}Various indoor scenes for mapping \\ and localization\end{tabular}                                   & \begin{tabular}[c]{@{}c@{}}Perception, Mapping, \\ Localization\end{tabular}                                                                                      \\ \rowcolor{C4!20}Euroc~\cite{euroc}                    & IJRR'16               & Indoor                      & \color{ForestGreen}{\cmark}       & \color{BrickRed}{\xmark}           & \color{BrickRed}{\xmark}         & \color{ForestGreen}{\cmark}      & \color{ForestGreen}{\cmark}              & \color{BrickRed}{\xmark}                 & \color{BrickRed}{\xmark}            & Motion Capture         & \begin{tabular}[c]{@{}c@{}}Various indoor scenes for mapping \\ and localization\end{tabular} & \begin{tabular}[c]{@{}c@{}}Perception, Mapping, \\ Localization\end{tabular} \\ 
\rowcolor{C4!20}ScanNet~\cite{scannet1}                  & CVPR'17               & Indoor                  &   \color{ForestGreen}{\cmark}          &  \color{ForestGreen}{\cmark}      &  \color{BrickRed}{\xmark}      & \color{BrickRed}{\xmark}     &  \color{ForestGreen}{\cmark}    &  \color{BrickRed}{\xmark}   & \color{BrickRed}{\xmark}                           & SLAM                       & \begin{tabular}[c]{@{}c@{}}A richly-annotated multiple room scenes \\ with semantic label\end{tabular}                          & \begin{tabular}[c]{@{}c@{}}Perception, Mapping, \\ Localization\end{tabular}                                                     \\
\rowcolor{C4!20}Replica~\cite{replica}                  & ArXiv'19              & Simulation Indoor       &  \color{ForestGreen}{\cmark}           &  \color{ForestGreen}{\cmark}      &  \color{BrickRed}{\xmark}       & \color{BrickRed}{\xmark}      &  \color{ForestGreen}{\cmark}    & \color{ForestGreen}{\cmark}    &  \color{ForestGreen}{\cmark}                         & Simulation                 & \begin{tabular}[c]{@{}c@{}}Highly photo-realistic 3D indoor scene \\ dataset at room scale\end{tabular}                         & \begin{tabular}[c]{@{}c@{}}Perception, Mapping, \\ Localization\end{tabular}                                                    \\
\rowcolor{C4!20}Nuscenes~\cite{nuscenes}                 & CVPR'20               & Outdoor                 &  \color{ForestGreen}{\cmark}           & \color{BrickRed}{\xmark}       &  \color{ForestGreen}{\cmark}      &  \color{ForestGreen}{\cmark}    &  \color{ForestGreen}{\cmark}    & \color{BrickRed}{\xmark}    &  \color{ForestGreen}{\cmark}                          & GNSS/INS                   & \begin{tabular}[c]{@{}c@{}}A city-scale dataset created for \\ autonomous driving research\end{tabular}                         & \begin{tabular}[c]{@{}c@{}}Perception, Mapping, \\ Localization\end{tabular}                                                   \\
\rowcolor{C4!20}Waymo~\cite{waymo}                   & CVPR'20               & Outdoor                 &       \color{ForestGreen}{\cmark}           & \color{BrickRed}{\xmark}       &  \color{ForestGreen}{\cmark}      &  \color{ForestGreen}{\cmark}    &  \color{ForestGreen}{\cmark}    & \color{BrickRed}{\xmark}    &  \color{ForestGreen}{\cmark}                          & GNSS/INS                   & \begin{tabular}[c]{@{}c@{}}A city-scale dataset created for \\ autonomous driving research\end{tabular}                         & Perception, Mapping                                                                   \\
\rowcolor{C4!20}Tartanair~\cite{tartanair}                   & IROS'20               & Outdoor                 &       \color{ForestGreen}{\cmark}           & \color{ForestGreen}{\cmark}       &  \color{ForestGreen}{\cmark}      &  \color{ForestGreen}{\cmark}    &  \color{ForestGreen}{\cmark}    & \color{ForestGreen}{\cmark}    &  \color{ForestGreen}{\cmark}                          & Simulation                   & \begin{tabular}[c]{@{}c@{}}A Large-scale dataset with various light conditions, \\ weather, and moving objects\end{tabular}                         & Perception, Mapping                                             \\
\rowcolor{C4!20}NTU VIRAL~\cite{ntu}                & IJRR'22               & Outdoor               &\color{ForestGreen}{\cmark}           & \color{BrickRed}{\xmark}       &  \color{ForestGreen}{\cmark}      &  \color{ForestGreen}{\cmark}    &  \color{ForestGreen}{\cmark}    & \color{BrickRed}{\xmark}    &  \color{BrickRed}{\xmark}             & TLS                        & A dataset of aerial mapping and localization                                                                                    & Mapping, Localization                                                                 \\
\rowcolor{C4!20}ScanNet++~\cite{scannet++}               & ICCV'23               & Indoor               &  \color{ForestGreen}{\cmark}          &  \color{ForestGreen}{\cmark}      &  \color{BrickRed}{\xmark}      & \color{BrickRed}{\xmark}     &  \color{ForestGreen}{\cmark}    &  \color{ForestGreen}{\cmark}   & \color{ForestGreen}{\cmark}          & SLAM                       & \begin{tabular}[c]{@{}c@{}}A large-scale dataset with high-quality and  \\ geometry and color of indoor scenes.\end{tabular}    & Perception, Mapping                                                                   \\ \rowcolor{C4!20}MARS-LVIG~\cite{mars}                    & IJRR'24               & Outdoor                      & \color{ForestGreen}{\cmark}       & \color{BrickRed}{\xmark}         & \color{ForestGreen}{\cmark}         & \color{ForestGreen}{\cmark}      & \color{ForestGreen}{\cmark}              & \color{BrickRed}{\xmark}                 & \color{BrickRed}{\xmark}            & GNSS/INS        & Aerial downward-looking  SLAM dataset & \begin{tabular}[c]{@{}c@{}}Perception, Mapping, \\ Localization\end{tabular} \\
  \rowcolor{C4!20}MCD~\cite{mcd}                      & CVPR'24               & Outdoor                 &  \color{ForestGreen}{\cmark}          &  \color{ForestGreen}{\cmark}      &  \color{ForestGreen}{\cmark}      & \color{ForestGreen}{\cmark}     &  \color{ForestGreen}{\cmark}    &  \color{ForestGreen}{\cmark}   & \color{ForestGreen}{\cmark}         & SMCTR          & \begin{tabular}[c]{@{}c@{}}A large-scale dataset with a wide\\  range of sensing modalities\end{tabular}                        & \begin{tabular}[c]{@{}c@{}}Perception, Mapping, \\ Localization\end{tabular}      \\  \rowcolor{C4!20}NES~\cite{mcnslam}                      & Arxiv'25               & Indoor/Outdoor                 &  \color{ForestGreen}{\cmark}          &  \color{ForestGreen}{\cmark}      &  \color{ForestGreen}{\cmark}      & \color{ForestGreen}{\cmark}     &  \color{ForestGreen}{\cmark}    &  \color{ForestGreen}{\cmark}   & \color{ForestGreen}{\cmark}         & SMCTR          & A large-scale indoor and outdoor 3D dataset                        & \begin{tabular}[c]{@{}c@{}}Perception, Mapping, \\ Localization\end{tabular}      \\                                     
\rowcolor{C3!20}RLBench~\cite{rlbench}               & RAL'20               & Indoor                  &    \color{ForestGreen}{\cmark}           &   \color{ForestGreen}{\cmark}      &  \color{BrickRed}{\xmark}      &  \color{BrickRed}{\xmark}    & \color{ForestGreen}{\cmark}     & \color{ForestGreen}{\cmark}    &      \color{ForestGreen}{\cmark}     & Simulation                        & \begin{tabular}[c]{@{}c@{}}large-scale learning environment featuring \\ 100 unique, hand-designed tasks.\end{tabular}                     & Manipulation \\
\rowcolor{C3!20}Robomimic~\cite{Robomimic}               & Arxiv'21               & Indoor                  &    \color{ForestGreen}{\cmark}           &   \color{BrickRed}{\xmark}      &  \color{BrickRed}{\xmark}      &  \color{BrickRed}{\xmark}    & \color{ForestGreen}{\cmark}     & \color{ForestGreen}{\cmark}    &      \color{BrickRed}{\xmark}     & Simulation                        & \begin{tabular}[c]{@{}c@{}}large-scale dataset consists of data collected from  \\ MachineGenerated , Proficient-Human, \\ and MultiHuman .\end{tabular}                     & Manipulation \\
\rowcolor{C3!20}Matterport3d~\cite{matterport3d}               & Arxiv'17               & Indoor                  &    \color{ForestGreen}{\cmark}           &   \color{ForestGreen}{\cmark}      &  \color{BrickRed}{\xmark}      &  \color{BrickRed}{\xmark}    & \color{ForestGreen}{\cmark}     & \color{ForestGreen}{\cmark}    &      \color{ForestGreen}{\cmark}    & Simulation                        & \begin{tabular}[c]{@{}c@{}}large-scale RGB-D dataset with 10,800 panoramic \\ views from 90 building-scale
scenes\end{tabular}                     & Navigation \\
\rowcolor{C3!20}HM3D~\cite{HM3D}               & CVPR'23               & Indoor                  &    \color{ForestGreen}{\cmark}           &   \color{ForestGreen}{\cmark}      &  \color{BrickRed}{\xmark}      &  \color{BrickRed}{\xmark}    & \color{ForestGreen}{\cmark}     & \color{ForestGreen}{\cmark}    &      \color{ForestGreen}{\cmark}    & Simulation                        & \begin{tabular}[c]{@{}c@{}}A object goal navigation dataset\end{tabular}                     & Navigation \\
\rowcolor{C7!20}Objaverse~\cite{objaverse}               & CVPR'23               & Indoor                  &    \color{ForestGreen}{\cmark}           &  \color{BrickRed}{\xmark}      &  \color{BrickRed}{\xmark}      &  \color{BrickRed}{\xmark}    & \color{ForestGreen}{\cmark}     & \color{ForestGreen}{\cmark}    &      \color{ForestGreen}{\cmark}     & CAD                        & \begin{tabular}[c]{@{}c@{}}A large dataset with descriptive captions, \\ tags, and animations.\end{tabular}                     & Generation                                                                            \\
\rowcolor{C7!20}OmniObject3D~\cite{omniobject3d}             & CVPR'23               & Indoor                  &       \color{ForestGreen}{\cmark}           &  \color{BrickRed}{\xmark}      &  \color{BrickRed}{\xmark}      &  \color{BrickRed}{\xmark}    & \color{ForestGreen}{\cmark}     & \color{ForestGreen}{\cmark}    &      \color{ForestGreen}{\cmark}      & CAD                        & \begin{tabular}[c]{@{}c@{}}Real-scanned 3D object captured \\ with both RGB and lidar\end{tabular}                              & \begin{tabular}[c]{@{}c@{}}Generation,  Perception, \\  Reconstruction\end{tabular} \\ \hline
\end{tabular}}}
\vspace{-15pt}
\end{table*}

\subsection{Geometric Representation Capability} In practical robotics applications like navigation and obstacle avoidance, geometric accuracy is prioritized over photorealistic appearance. 
Traditional scene representations exhibit strong capability in modeling scene geometry. Dense point cloud representations can effectively capture fine geometric details of a scene, but they often lead to excessive memory consumption. Voxel-based representations can alleviate memory usage to some extent while preserving geometric accuracy. In contrast, neural scene representations rely on network parameters and high-level features to model scenes, which generally results in weaker geometric fidelity.

\noindent\textbf{\textit{Geometric Representations}}
PointNet~\cite{pointnet} proposes a novel neural network architecture with shared multi-layer perceptrons (MLPs) and max pooling to aggregate global features for 3D
recognition task. PointNet++~\cite{pointnet++} extends the original PointNet by incorporating hierarchical learning, allowing the network to capture
local structures at multiple scales. 
3D ShapeNets~\cite{shapenet} applies deep belief networks (DBNs) to voxel representations of 3D shapes.  DeepSDF~\cite{deepsdf} introduces an SDF-based scene representation for modeling continuous geometry. It proposes an autoencoder--decoder framework that learns implicit signed distance functions to represent smooth and detailed object surfaces. In parallel, a line of work~\cite{deepmesh} explore mapping 3D shape surfaces onto 2D geometry images, enabling the direct use of conventional 2D CNNs for shape analysis and reconstruction. 

\noindent\textbf{\textit{NeRF-Based  Representations}}
NeRF use volumetric density formulation to represent the scene often leads to blurry or inaccurate geometry. To address this, UNISURF~\cite{UNISURF} unifies neural implicit surfaces and radiance fields, enabling multi-view surface reconstruction without foreground masks. NeuS~\cite{NEUS} introduces a signed distance function (SDF) parameterization together with a novel volume rendering formulation, yielding high-fidelity surfaces while retaining photorealistic rendering. Similarly, VolSDF~\cite{volsdf} models density as a transformed SDF and develops adaptive sampling with provable error bounds, improving surface accuracy.

\noindent\textbf{\textit{3DGS-Based  Representations}}
 The original 3DGS method lacks geometric supervision, resulting in relatively weak geometric representation of the scene. Recently, several works have been proposed to improve the geometric accuracy of the 3D reconstruction. SuGaR~\cite{guedon2024cvpr-sugar} uses a hybrid representation to align the Gaussians with the surface as an early attempt in this direction. Since the 3D Gaussian representation has the inevitable ambiguity in surface description, a natural solution is to flatten the 3D Gaussians as proposed in 2D Gaussian Splatting (2DGS)~\cite{huang2024siggraph-2dgs} and Gaussian Surfels~\cite{gaussiansurfel}. With the disk-like 2D surfels in the 3D space, one can render depth and normal accurately.  Another line of work combines Gaussian Splatting and volume rendering for better surface reconstruction. Gaussian opacity field~\cite{gaussianopcity} models the discrete explicit 3D Gaussian representation as an opacity field via volume rendering. Similar to the distance field, surface meshes can then be extracted from the opacity field by the marching cubes algorithm. 

\noindent\textbf{\textit{Foundation Model}}
Token-based scene representations often suffer from limited geometric expressiveness, which restricts their ability to accurately capture 3D structure. To address this issue, several approaches explicitly incorporate geometric priors to enhance representation ability. GPNR~\cite{gpnr} integrates epipolar geometry into its encoder--decoder architecture to enforce multi-view consistency, while \cite{stereopair} propose a multiview vision transformer with epipolar line sampling to improve geometry reconstruction from images. GTA~\cite{gta} develops geometric attention to embed 3D information into tokens, which is then combined with Scene Representation Transformers (SRT) to enhance scene reconstruction. Beyond scene-level transformers, a complementary line of work focuses on mesh tokenization for 3D generation. MeshGPT~\cite{meshgpt} employs a VQ-VAE network to learn a discrete mesh vocabulary and autoregressively generates triangle meshes via a transformer decoder. To mitigate cumulative errors inherent in VQ-VAE sequence modeling, MeshXL~\cite{meshxl} leverages a neural coordinate field for sequential mesh representation, enabling higher-quality autoregressive mesh synthesis.  Collectively, these advances highlight a growing trend of combining geometric reasoning with token-based models to improve scene representation and 3D generation quality.
\begin{figure}[t]
    \centering
    \includegraphics[width=\linewidth]{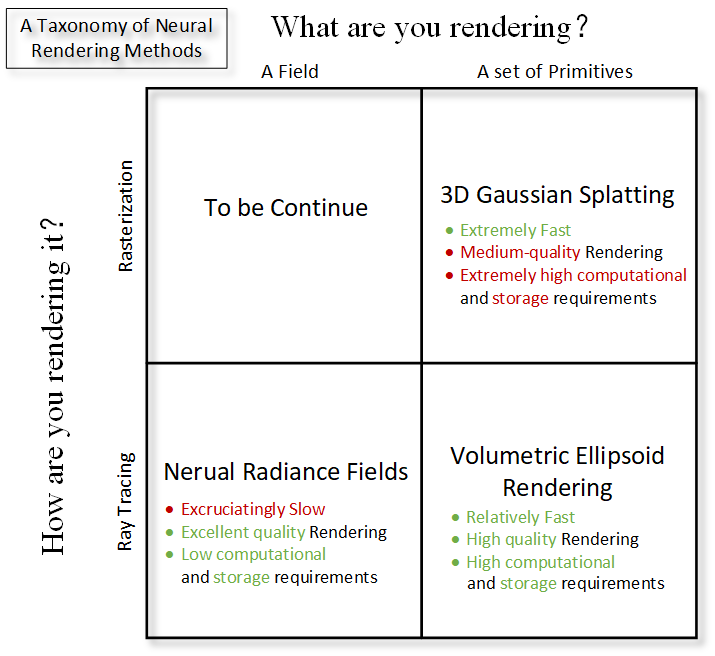}
    \vspace{-15pt}
    \caption{ A Taxonomy and future work of neural scene representation: NeRF\cite{NeRF}, 3DGS~\cite{3dgs} and volumetric ellipsoids rendering~\cite{ever}.
}
    \label{fig:nerf3dgs2}
    \vspace{-15pt}
\end{figure}
% \noindent\textbf{Future work and Comparison with NeRF}

% Accurate representation of real-world geometry and physical properties is critical for the reliable operation of robotics systems. Future research directions primarily include: (a) enhancing geometric modeling capabilities by improving the Gaussian-based scene representation, and (b) incorporating physical priors and multi-modal features in a more effective and integrated manner. 

% Overall, 3DGS-Based physical scene understanding holds significant potential due to its explicit and interpretable nature. However, in terms of precise surface geometry modeling---especially for planar structures---the inherent shape characteristics of Gaussian ellipsoids may lead to limitations, rendering them less effective compared to implicit representations like NeRF.

\section{Perception}
\label{sec:perception}

Perception in robotics bridges the gap between raw sensory inputs and actionable knowledge. While geometric representations capture spatial structure, true autonomy requires grounding these representations with semantic meaning. This enables robots not only to perceive objects as geometric shapes but also to recognize their identity, attributes, and contextual roles within a scene.
In this section, we divide semantic grounding into two complementary levels: (i) object-centric semantic representations, which emphasize the recognition and representation of individual objects, and (ii) scene-level semantic understanding, which focuses on the holistic interpretation of scenes, relationships among entities, and task-driven reasoning.
This layered perspective reflects the dual needs of robotics: precise recognition of objects for manipulation and navigation, and broader semantic comprehension for planning and decision-making in complex environments.

\subsection{Object Detection}
% Object-centric semantic representations focus on encoding the semantic meaning of individual objects within the environment.
% This representation is crucial for object classification, detection, segmentation, and affordance reasoning.
% We review existing approaches across seven paradigms:
Object-centric representations focus on modeling and identifying individual objects within a scene, serving as the foundation for object-level perception in robotics.
Such representations are crucial for object detection, classification, and affordance reasoning, enabling robots to localize and interpret discrete entities for manipulation and navigation.
In this subsection, we review representative methods across different 3D representation paradigms.

\noindent\textbf{\textit{Geometric Representations}}
% Traditional semantic grounding methods rely on explicit geometric representations such as points, voxels, and signed distance fields. Point-based approaches~\cite{lai2025natural} directly operate on LiDAR or RGB-D point sets and form one of the earliest paradigms, enabling lightweight and real-time recognition and object-level planning with methods such as PointNetGPD~\cite{qi2019pointnetgpd}. Voxel grids provide dense, structured representations of 3D space~\cite{fan2025structured}, with VoxelNet~\cite{zhou2018voxelnet} and SparseConvNet~\cite{graham20183d} pioneering end-to-end recognition, while MinkowskiNet~\cite{choy20194d} scales sparse convolutions for high-resolution semantic tasks. Signed Distance Fields (SDFs) represent smooth surfaces implicitly~\cite{deepsdf} and support robust mapping. While effective, these geometric encodings share key limitations: point clouds are sparse and occlusion-prone, voxels are memory-intensive and resolution-limited, and SDFs require volumetric storage with limited scalability for open-vocabulary grounding, motivating the shift toward neural and hybrid scene representations.
Traditional object detection frameworks in 3D perception primarily rely on explicit geometric representations such as point clouds and voxels.
Point-based approaches directly operate on LiDAR or RGB-D point sets and remain among the earliest paradigms for 3D object detection.
Representative works include MVPDet \cite{yin2021mvpdet}, which fuses cross-modal geometric and visual cues through virtual sampling, and PIXOR \cite{yang2018pixor}, which achieves efficient BEV-style detection using raw LiDAR data.
Further developments such as 3DSSD \cite{yang20203dssd} and CenterPoint \cite{yin2021centerpoint} improve feature aggregation and anchor-free prediction, while maintaining real-time performance for large-scale autonomous driving scenes.
Voxel-based approaches provide structured volumetric representations that enable efficient convolutional processing.
VoxelNet \cite{zhou2018voxelnet} first introduced an end-to-end voxel encoding pipeline, where point features within each voxel are aggregated via PointNet-style layers.
Later extensions such as PointPillars \cite{lang2019pointpillars} simplified voxelization into vertical pillars for real-time deployment.
Although these explicit encodings achieve reliable performance in structured environments, they exhibit inherent limitations: point clouds are sparse and occlusion-prone, while voxel grids are memory-intensive and limited by discretization.
As a result, recent research trends have shifted toward implicit or neural scene representations, such as NeRF and 3D Gaussian Splatting, which offer improved continuity and richer geometric--appearance coupling.

\noindent\textbf{\textit{NeRF-Based Representations}}
%NeRF~\cite{NeRF} implicitly model 3D scenes as continuous volumetric functions mapping spatial coordinates and viewing directions to color and density. Extensions such as Semantic-NeRF~\cite{zhi2021semanticnerf} jointly reconstruct geometry, appearance, and per-voxel semantics, enabling fine-grained semantic recovery from multi-view images.Recently, NeRF representations have been integrated into 3D detection pipelines: NeRF-Det~\cite{xu2023nerfdet} learns geometry-aware volumetric features for multi-view 3D detection, NeRF-Det++~\cite{Huang2025NeRFDetpp} enhances it with semantic cues and perspective-aware depth supervision, and GO-N3RDet~\cite{li2025gon3rdet} further optimizes geometry through position-encoded occupancy modeling and importance sampling.These advances demonstrate how NeRF can serve as a geometry-consistent backbone for joint reconstruction and object-level perception. However, NeRF-Based detectors remain computationally intensive and view-dependent, limiting their scalability to large or dynamic scenes and making explicit boundary encoding and lightweight fine-tuning difficult.
NeRF represent 3D scenes as continuous volumetric functions that map spatial coordinates and viewing directions to color and density.
Recent works have leveraged this implicit 3D formulation to integrate geometric consistency into object detection.
\textit{NeRF-Det}~\cite{xu2023nerfdet} introduces a geometry-aware volumetric backbone that learns dense 3D features from multi-view images for end-to-end 3D object detection.
\textit{GO-N3RDet}~\cite{li2025gon3rdet} further enhances this design through geometry-optimized occupancy modeling and importance sampling, improving detection precision while reducing sampling redundancy.
Together, these methods demonstrate that NeRF can serve as a geometry-consistent and differentiable backbone for multi-view 3D object detection.
However, NeRF-Based detectors remain computationally demanding and sensitive to viewpoint distribution, limiting their scalability in dynamic or large-scale environments.

\noindent\textbf{\textit{3DGS-Based Representations}}
% 3D Gaussian Splatting (3DGS) represents scenes with collections of anisotropic Gaussians that can be rendered 
% in real time~\cite{kerbl2023gaussian}. At the object level, semantic extensions such as SaGA~\cite{cen2023saga} 
% and OpenGaussian~\cite{wu2024opengaussian} enrich Gaussian splats with segmentation priors or CLIP features, 
% enabling open-vocabulary recognition and cross-modal queries for individual objects. These methods~\cite{qi2024air} provide 
% efficient, photorealistic reconstructions with semantic grounding, making them suitable for object detection, 
% classification, and manipulation. However, while they offer significant speed advantages over NeRF, their 
% semantic quality is still limited by 2D priors and sparse training views, constraining robustness in cluttered 
% or occluded object settings.
3D Gaussian Splatting (3DGS) provides an explicit and differentiable representation of scenes through a collection of anisotropic Gaussian primitives, enabling efficient rendering and geometry reasoning.
Recent works have begun to leverage 3DGS for object-level detection and pose estimation.
\textit{6DGS}~\cite{6dgs2024eccv} formulates single-image 6D object pose estimation as a correspondence problem between image features and a prebuilt 3D Gaussian model, achieving accurate localization and orientation recovery.
\textit{MATT-GS}~\cite{mattgs2025arxiv} introduces a masked-attention Gaussian backbone for robotic object detection, enhancing robustness under occlusions and lighting variations.
\textit{GS2Pose}~\cite{gs2pose2024arxiv} further integrates a two-stage refinement mechanism guided by Gaussian-rendered geometry, bridging detection and 6D pose estimation.
Together, these approaches demonstrate that Gaussian splatting can serve as a compact and geometry-consistent representation for object-level reasoning; however, most current works remain limited to small-scale scenes or specific object categories due to the high computational cost of dense Gaussian optimization.

% \noindent\textbf{Token-Based Representations}
% Token-based methods discretize 3D geometry into compact tokens processed by large-scale transformers, enabling object-level semantic grounding through open-vocabulary reasoning and zero-shot transfer~\cite{huang2024chat}. For example, QueryGS~\cite{wang2024query} embeds CLIP-guided semantic tokens into Gaussian splats, supporting flexible object queries and integration with reinforcement learning. Such approaches provide a lightweight bridge between 3D geometry and language models, but remain sensitive to tokenization quality and have yet to fully match the geometric fidelity of continuous representations.
% \noindent\textbf{Token-Based Representation.}
% Object-centric tokenization represents each 3D object with compact tokens or embeddings processed by large transformers, enabling zero-shot recognition and language grounding. Representative approaches include Point-BERT~\cite{Yu2022PointBERT}, which discretizes local point patches into a codebook of tokens for masked pre-training, PointCLIP~\cite{Zhang2022PointCLIP}, which aligns object-level point clouds to CLIP text--image tokens via multi-view projections, and ULIP~\cite{Xue2023ULIP} and ULIP-2~\cite{Xue2024ULIP2}, which pre-train on object triplets (point cloud--image--text) to learn unified, transferable object embeddings. Despite their efficiency and transferability, these object tokens remain sensitive to tokenization and alignment quality, and still trail continuous fields in fine geometric fidelity.

\noindent \textit{\textbf{Foundation Model}}
Recent vision-language foundation models have extended traditional object detectors toward open-vocabulary and zero-shot perception.
Grounding DINO~\cite{liu2023groundingdino} combines DINO's visual encoder with CLIP-based textual embeddings to perform prompt-driven detection without category-specific retraining.
YOLO-World~\cite{yoloworld2024cvpr} incorporates vision-language pre-training into the YOLO architecture, enabling real-time open-vocabulary detection with high inference efficiency.
YOLO-E~\cite{yoloe2024arxiv} further improves cross-modal alignment by introducing a unified vision-language encoder and a lightweight grounding head, achieving scalable and efficient detection under free-form text prompts.
These models demonstrate that foundation-model priors can generalize object detection beyond closed-category datasets, effectively bridging perception and natural language grounding for robotics and embodied AI applications.

\subsection{Scene-Level Semantic Understanding}
Scene-level semantic understanding emphasizes holistic reasoning about the environment, including object relationships,
scene graphs, panoptic representations, and task-oriented semantics. We review existing approaches across six paradigms:

\noindent\textbf{\textit{Geometric Representations}}
Scene-level methods have long relied on explicit geometric representations such as points, voxels, and signed distance fields.
Point-based approaches construct scene graphs or contextual embeddings directly from point clouds~\cite{wu2021scenegraphfusion,qi2021relationshipgraph}, enabling reasoning over spatial relations for grounding and task planning.
Recent advances such as PointContrast~\cite{xie2020pointcontrast} and the unified query-based paradigm for point cloud understanding~\cite{wang2024unifiedquery} introduce large-scale pre-training and transformer-based querying for richer semantic scene understanding, improving robustness under sparse or partial observations.
{Beyond these, hierarchical 3D scene graphs unify semantics and geometry in a queryable structure~\cite{armeni20193d}, with real-time construction in robotics~\cite{hughes2022hydra}, while transformer-based point-cloud methods directly infer 3D scene graphs and relations~\cite{lv2024sgformer}.}
Voxel grids, in turn, provide structured volumetric maps for dense completion and semantics.
Classical systems like OctoMap~\cite{octomap} maintain probabilistic occupancy for long-term exploration, while modern learning-based methods such as VoxFormer~\cite{li2023voxformer} and VMNet~\cite{hu2022vmnet} leverage sparse voxel transformers and voxel--mesh coupling to achieve high-fidelity 3D semantic scene completion and segmentation.
Signed Distance Fields (SDFs) further unify geometry, semantics, and topology, as in 
Recent works like SurroundSDF~\cite{jiang2023surroundsdf} extend implicit SDF representations to capture semantic structure in complex indoor and driving scenes, enabling continuous field reasoning with improved surface consistency.
Together, these representations bridge geometric mapping and semantic abstraction, laying the foundation for large-scale, language-grounded 3D scene understanding.

\noindent\textbf{\textit{NeRF-Based Representations}}
For scene-level understanding, NeRFs provide a powerful implicit representation that captures continuous geometry and appearance across entire environments, supporting globally consistent scene reconstruction and reasoning. Beyond pure geometry, semantic extensions incorporate label supervision to recover per-pixel semantics~\cite{snislam}, and 
language-conditioned variants such as CLIP-NeRF~\cite{wang2022clipnerf} and LERF~\cite{kerr2023lerf} embed open-vocabulary features into the radiance field, enabling zero-shot recognition, natural language queries, and task grounding at the scene scale. While these approaches unify geometry, appearance, and semantics in a 
single model, they remain computationally expensive and require dense multi-view training, limiting their scalability to large or dynamic environments and constraining deployment in real-time robotics.

\noindent\textbf{\textit{3DGS-Based Representations}}
At the scene level, 3DGS-Based methods extend beyond object recognition to holistic mapping and navigation. Approaches such as GaussianBeV~\cite{chabot2024gaussianbev} adapt Gaussian splats for bird's-eye-view segmentation, enabling large-scale scene understanding in autonomous driving and robotic navigation. By combining explicit geometry with language-grounded semantics, these methods support open-vocabulary queries and task-level reasoning across entire environments. While 3DGS achieves real-time efficiency and scalability, its reliance on view-based supervision and limited integration with long-horizon planning currently restrict its adoption as a comprehensive scene representation in robotics.

%\noindent\textbf{Token-Based Representations.}
\noindent\textbf{\textit{Foundation Model}}
At the scene level, token-based representations extend beyond individual objects to treat entire regions and relations as tokens that can be reasoned about via language models.
Scene-LLM~\cite{li2023scenellm} grounds large language models directly in 3D scenes, allowing robots to query, plan, and interpret environments using natural language.
Complementary approaches such as NLMap~\cite{nlmap} build open-vocabulary queryable scene maps for real-world planning,
while CLIP-Fields~\cite{clipfields2023} embed language features into NeRF-Based radiance fields for text-driven scene understanding.
This paradigm enables seamless integration of perception with high-level reasoning, facilitating context-aware planning and human-robot interaction.
However, while promising, current methods struggle with scalability and maintaining geometric accuracy at large scales, limiting their immediate use as standalone scene representations.

Across object-centric and scene-level perspectives, scene representations span from explicit geometry to implicit neural fields and emerging tokenized models.
Object-level methods enable precise recognition and affordance reasoning, whereas scene-level approaches capture context and relations for planning and language grounding.
Geometric encodings remain efficient but limited; neural fields add rich semantics at high cost; and newer paradigms promise open-vocabulary grounding yet still lack robustness---motivating hybrid designs that balance fidelity, scalability, and generalizability.

\section{Mapping \& Localization}
\label{sec:mapping}

\begin{figure*}[t]
    \centering
    \includegraphics[width=\linewidth]{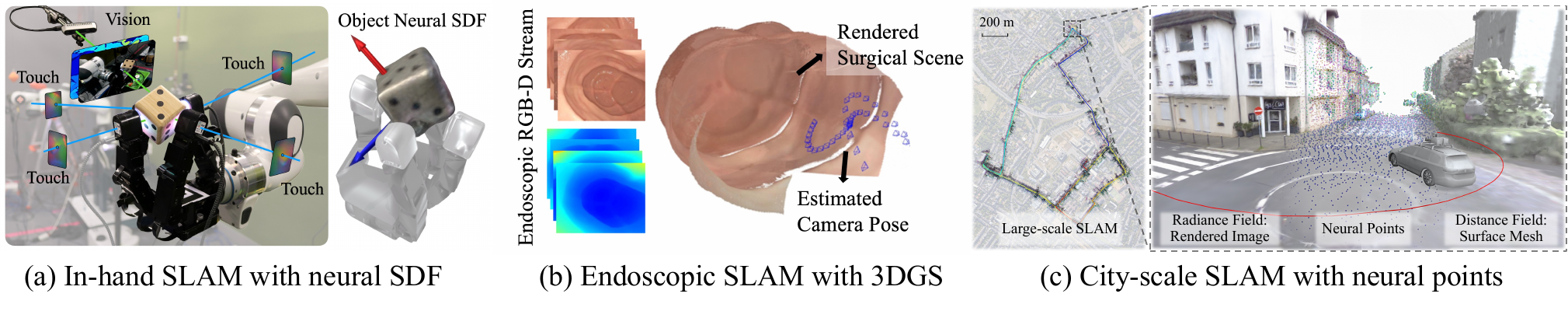}
    \vspace{-18pt}
    \caption{Mapping and localization using different scene representations for applications at different scales: (a)~\cite{suresh2024neuralfeels}, (b)~\cite{huang2025icra-endo2dtam}, and (c)~\cite{pan2025pings}
}
\label{fig:slam_example}
\vspace{-15pt}
\end{figure*}

\subsection{Scene Reconstruction}

Dense mapping reconstructs detailed geometry once the trajectory (or an estimate) is available, exploiting full sensor data beyond sparse landmarks. Practical systems balance what to estimate (occupancy, distance fields, surfaces), how to represent it (explicit points/surfels/meshes vs. implicit fields), and which data structures to use (voxels, octrees, hashed grids, hierarchical trees). The choice depends on downstream tasks: navigation favors occupancy or ESDF for collision checks; manipulation/AR often prefers meshes/surfels/3DGS for editability; inspection benefits from high-resolution explicit surfaces.

\noindent\textbf{\textit{Geometric  Representations}}
 Classical multi-view reconstruction pipelines recover scenes as sparse or dense point sets from calibrated images and depth sensors. COLMAP-style SfM/MVS pipelines~\cite{schoenberger2016sfm} provide accurate geometry but require extensive viewpoints and struggle in texture-less or reflective regions. Point sets are typically post-processed into meshes via surface reconstruction, e.g., Poisson or Delaunay variants, yielding watertight models for rendering and simulation~\cite{mesh,mesh2}. Point-based approaches are interpretable and easy to fuse using range sensors such as LiDAR and depth cameras, but their storage grows with area coverage and they provide no native view-dependent appearance.
Voxel grids discretize space and store occupancy, signed distance, or radiance features. In robotics, OctoMap~\cite{octomap} scales occupancy mapping through octrees, explicitly modeling free, occupied, and unknown space, while ESDF/TSDF systems like Voxblox~\cite{voxblox,pan2022iros} support planning and surface extraction with real-time updates. For photorealistic view synthesis, explicit radiance voxels and factorized variants trade memory for speed and parallelism, and integrate well with local mapping. Voxelization enables efficient proximity indexing and convolutional processing but suffers from resolution--memory trade-offs on large scenes. 
In addition to voxelized SDF, neural SDF can encode scenes as continuous implicit functions over space. DeepSDF~\cite{deepsdf} learns object-level shape priors for high-fidelity surface recovery and can be extended to the scene level~\cite{ortiz2022isdf}. SDFs provide excellent geometry for manipulation and contact reasoning, while requiring either dense sampling or acceleration structures for efficient querying.

\noindent\textbf{\textit{NeRF-Based Representations}}
NeRF~\cite{NeRF} represents scenes as continuous volumetric fields of density and color, enabling both high-quality novel view synthesis and geometry reconstruction from posed images. UniSURF~\cite{UNISURF} presents a unified view of implicit surfaces and radiance fields to improve multi-view surface reconstruction while maintaining view synthesis quality. Neus~\cite{NEUS} first introduced a volume integration loss for an unbiased surface representation of the volume density. Neuralangelo~\cite{li2023neuralangelo} improves reconstruction scalability and fidelity by introducing multi-resolution voxel hashing as in~\cite{instantngp} and the numerical calculation of the SDF gradient. Several works~\cite{tancik2022blocknerf,tao2024silvr,prosgnerf} use a divide-and-conquer strategy to reconstruct even larger scenes with NeRF.
% VolSDF~\cite{volsdf} improves geometric accuracy and produces clean, watertight meshes via Marching Cubes.

% Subsequent works improve anti-aliasing and unbounded scene handling (Mip-NeRF 360~\cite{mipnerf360}), and reduce memory via factorized or grid-encoded fields (e.g., TensoRF~\cite{tensorf}, Plenoxels~\cite{plenoxels}). NeRF-style methods deliver photorealism and implicit regularization but can be compute-intensive at training and rendering time; 

\noindent\textbf{\textit{3DGS-Based Representations}}
3DGS~\cite{3dgs} models scenes with lots of anisotropic Gaussians optimized for real-time rasterization. 
Compared to NeRF, 3DGS dramatically accelerates training and rendering and offers explicit, editable primitives. One limitation of the naive 3DGS is the geometry ambiguity when rendering the depth or surface normal. There is also no direct conversion from 3DGS to a dense surface representation, such as SDF and mesh.
Several methods solve this by either using 2D Gaussian surfels instead of 3D ellipsoids~\cite{2dgs,gaussiansurfel,radegs} or adopting hybrid opacity or distance field formulations~\cite{gaussianopcity}. 

% Because splatting treats pixels as points, vanilla 3DGS can exhibit aliasing across scales. Mip-splatting~\cite{mipsplat}, EWA filtering~\cite{ewa}, and Analytic-Splatting~\cite{analyticsplatting} mitigate blurring/jaggies by integrating a Gaussian signal over pixel footprints, delivering high-quality zoom-out and level-of-detail (LOD) behavior and improving cross-resolution view synthesis. GaussianPro~\cite{gaussianpro} leverages SfM priors and progressive densification to produce better-initialized primitives, improving geometric convergence, especially under challenging viewpoints.
% However, naive models can be memory-hungry without compression or hierarchical structure. 

To further deploy 3DGS at scale for photorealistic and geometrically accurate scene reconstruction, numerous works attempt to reduce 3DGS memory~\cite{contextgs,scaffoldgs} or employ hierarchical data structures~\cite{octreegs,kerbl2024hierarchical3dgs,liu2024citygaussianv2}. Scaffold-GS~\cite{scaffoldgs} organizes local anchors in regular voxel grids and spawns Gaussian primitives from the anchor features with shared MLPs, while OctreeGS~\cite{octreegs} uses an octree structure to further decrease memory consumption. For aerial/urban scenes, hierarchical 3DGS~\cite{kerbl2024hierarchical3dgs}, CityGaussian~\cite{liu2024citygaussianv2} and PINGS~\cite{pan2025pings} demonstrate real-time city-scale rendering, while driving-scene pipelines~\cite{zhou2024drivinggaussian, yan2024eccv-streetgs} reconstruct high-fidelity radiance with explicit modeling of dynamic objects. 

\noindent\textbf{\textit{Foundation Model}}
The emergence of visual token-based feed-forward models has opened a new path: instead of reconstructing scenes through hand-crafted geometric solvers, scenes are represented through sets of visual tokens learned from large-scale data, enabling direct prediction of geometry, appearance, and pose.

In the area of mapping for 3D reconstruction, pointmap-based approaches such as DuSt3R~\cite{wang2024cvpr-dust3r} demonstrated that two-view transformers can directly regress dense pointmaps represented internally as visual tokens exchanged across views. This abstraction provides both explicit scene geometry and dense correspondences, allowing downstream pose estimation and mapping without iterative triangulation.
DuSt3R was soon extended by a couple of works that adapt this framework to larger collections and streaming settings. MASt3R~\cite{leroy2024eccv} improves multi-view consistency by introducing global reasoning over many image tokens, while Spann3r~\cite{wang2025threedv} and MUSt3R~\cite{cabon2025cvpr} scale the architecture to online SLAM by equipping it with smart memory and retrieval mechanisms. Similarly, a series of works~\cite{yang2025cvpr, tang2025cvpr, wang2025cvpr-vggt} treat geometry estimation itself as a token prediction problem, where 3D structure is encoded as a set of latent tokens grounded in visual input. By scaling the capacity of the model and the size of the training data, and using the alternating attention architecture, the seminal work VGGT~\cite{wang2025cvpr-vggt} manages to reconstruct scenes for a larger batch of images and achieves a better performance compared to previous models. This pipeline is then followed by models such as Pi-3~\cite{wang2025arxiv-pi3}, which further remove the need for a reference frame. Recently, similar to Pow3r~\cite{jang2025cvpr} for the Dust3r model, MapAnything~\cite{keetha2025arxiv-mapanything} adds additional prior information such as known camera parameters and depth measurements into the VGGT architecture to further realize a universal mapping model.
Some other works~\cite{zhang2025monst3r,li2025megasam,xiao2025spatialtracker} focus more on the modeling and tracking of dynamic objects by adding additional tracking heads and using more data from dynamic scenes during training. 

% In parallel, for the novel view synthesis task, a rich taxonomy of approaches has emerged depending on the 3D representation. Early feed-forward NeRF-style models, such as PixelNeRF~\cite{yu2021pixelnerf}, conditioned neural radiance fields on multi-view image tokens to enable novel view generation without per-scene optimization. While photorealistic, these methods remained computationally demanding. Subsequent paradigms~\cite{charatan2024pixelsplat, ye2025iclr}  directly predict pixel-aligned rasterizable 3D Gaussian primitives, dramatically improving rendering speed via splatting while maintaining quality. This is then extended from pairwise image input or sparse views to a collection or stream of image inputs~\cite{zhang2025flare, jiang2025anysplat} following the VGGT scheme. LVSM~\cite{jin2025lvsm}, on the other hand, eliminates the inductive bias of specific 3D representations and directly generates high-quality novel view images. 

% Mention also Monst3r (and some other works, Spatialtracker, MegaSAM) for handling the dynamic objects

% Mention also the works based on large reconstruction model (like LVSM)

Collectively, these methods exemplify a new design principle: represent a scene not as sparse hand-engineered features but as dense sets of learned visual tokens from which dense geometry can be recovered.
The transition to token-based models has been enabled by scaling laws in 3D vision. Such systems rely on large model capacities and are trained on massive multi-view datasets. The effectiveness of these approaches validates the hypothesis that, as in natural language processing, performance in 3D reconstruction improves predictably with model and data scale.

\subsection{SLAM}

\noindent\textbf{\textit{Geometric Representations}}
Feature-based visual SLAM~\cite{vins,orbslam2} and LiDAR odometry~\cite{kissicp,lou2025qlio,cte-mlo,magmm} estimate poses by matching sparse points or registering scans, with maps as keypoints, surfels, or BEV map. For dense mapping with RGB-D and LiDAR, surfel maps~\cite{elasticfusion,suma} can maintain local surface elements with normals, colors, and confidence, supporting fast rendering-based tracking and post-hoc deformation graphs for loop-closure corrections. These pipelines are robust and efficient, but produce sparse or piecewise-dense geometry and offer limited photorealistic rendering for downstream perception or teleoperation. 

Volumetric occupancy maps can support dense tracking and online fusion. The early work for 2D LiDAR SLAM~\cite{dellaert1999monte, kohlbrecher2011hector} is mainly based on the occupancy grid map. Driven by the probabilistic robotics theory, the occupancy probability stored in grids is incorporated into a particle filter for pose tracking and map update. 
Although tracking using discrete volumetric SDF maps has already been applied in classic systems such as KinectFusion~\cite{kinectfusion} and Voxblox~\cite{voxblox}, the use of implicit neural SDF maps is gaining increasing attention for localization due to their continuity and differentiability.  
%%%%补充一两篇scene graph的论文

Some recent methods adopt scene graph--based representations to fuse geometry, semantics, and spatial relations into structured and queryable maps. Hydra~\cite{chang2023hydra} builds multi-layer 3D scene graphs at the level of objects, rooms, and buildings. They integrate geometric reconstruction and semantic reasoning for consistent large-scale mapping. ConceptGraphs~\cite{gu2024conceptgraphs} fuses visual detections, depth, and pose estimates into a persistent 3D object-level map represented as a scene graph encoding spatial and semantic relationships among entities.

\noindent\textbf{\textit{NeRF-Based Representations}}
Both NeRF and 3DGS are scene representations that support differentiable volume rendering. To minimize the difference bewteen the rendered images and the actual observed images, one can either optimize the parameters of the scene representation as already mentioned above or the camera poses as we will discuss in this section.
NeRF-Based SLAM jointly optimizes camera poses and radiance fields from streaming images. Systems such as iMap~\cite{imap} leverage differentiable rendering with MLP-based NeRF for rendering-based tracking, while maintaining dense photorealistic maps. Later, Nice-SLAM~\cite{niceslam}, ESLAM~\cite{eslam}, and other works~\cite{plgslam,coslam,pointslam,neslam, ddnslam} make use of a hybrid representation combining the local optimizable features stored in various explicit spatial data structures or multi-submap scene representation to enhance the map scale and fidelity, as well as the tracking accuracy.   MNE-SLAM~\cite{deng2025mne} and MCN-SLAM~\cite{mcnslam} propose the distributed multi-agent collaborative SLAM framework with distributed mapping and camera tracking, joint scene representation, intra-to-inter loop closure, and multi-submap fusion.  NeuralFeels~\cite{suresh2024neuralfeels} proposes a multimodal perception method that integrates vision and touch to improve spatial awareness and object tracking during hand manipulation with a neural distance field. Some LiDAR SLAM systems~\cite{nerfloam,pinslam} also make use of the neural distance field with design logic similar to the aforementioned NeRF SLAM systems and scale well to larger outdoor scenes.
% provides superior surfaces for planning and interaction, at the cost of higher memory and compute than sparse pipelines. 

\noindent\textbf{\textit{3DGS-Based Representations}}
3DGS has enabled real-time, map-centric SLAM with explicit, differentiable primitives and fast rasterization. Decoupled systems use external odometry (e.g., Photo-SLAM~\cite{huang2024cvpr-photoslam}, GS-ICP-SLAM~\cite{ha2024eccv}, while optimizing Gaussians in the back-end. Coupled systems (SplaTAM~\cite{keetha2024cvpr-splatam}, MonoGS~\cite{matsuki2024cvpr-monogs}, and so on~\cite{densesplat,mgslam, sgsslam}) track directly against the Gaussian map via rendering losses, with recent work adding loop closure (LoopSplat~\cite{zhu2025threedv-loopsplat}, VPGS-SLAM~\cite{vpgs}). These approaches provide dense geometry, photorealistic rendering, and editable maps with superior runtime compared to NeRF-Based SLAM. For pose optimization, the direct differentiation through per-primitive transformations is accurate but costly and may violate the \(\mathrm{SE}(3)\) manifold constraint. MonoGS~\cite{matsuki2024cvpr-monogs} derive analytical Jacobians on \(\mathrm{SE}(3)\) within a CUDA rasterizer, improving stability and speed. Recently, Gaussian Splatting-based radiance fields have been applied as the map representation for endoscopic SLAM~\cite{Wang2024EndoGSLAM} to provide the doctors with photorealistic rendering of the organs during the operations.

\noindent\textbf{\textit{Foundation Model}}
Visual SLAM has long followed a geometry-driven pipeline (feature detection/matching, motion estimation, bundle adjustment, explicit mapping). Despite the learned features~\cite{superpoint} and optical flow~\cite{droidslam} improve individual modules, the pipeline remains brittle and fragmented. The pioneer work DuSt3R~\cite{wang2024cvpr-dust3r} takes a radically different path. Instead of refining one module, it replaces the entire chain with a single vision transfomer-based model that maps pairs of images directly to dense 3D structures in the form of a pixel-aligned point map, from which the camera parameters can also be recovered. The follow-up work MASt3R~\cite{leroy2024eccv} adds a pixelwsie feature matching head with explicit supervision to further improve the performance while VGGT~\cite{wang2025cvpr-vggt} manages to go beyond the pairwise setting and reconstruct a couple of images in one shot.

Several works incorporate results from feedfoward 3D vision foundation models into different submodules of the SLAM pipeline. For instance, MASt3R-SLAM~\cite{murai2025cvpr-mast3rslam} exploits feature matching results from MASt3R as the front-end for back-end bundle adjustment while VGGT-SLAM~\cite{maggio2025neurips} merges submaps from different VGGT processing batches by scale-aware pose graph optimization to bypass the memory bottleneck of processing thousands of images. Following the design of previous learning-based visual SLAM using DBA~\cite{droidslam}, ViPE~\cite{huang2025arxiv-vipe} trains its own model to predict depth and optical flow for front-end tracking and back-end optimization, with attention to dynamic objects in the scene. By leveraging strong priors from these freezed foundation models, these works achieve a robust online SLAM pipeline, even without the need for known camera intrinsics.

Another line of work is modeling the SLAM problem more directly with token-based vision models. For example, SLAM3R~\cite{liu2025cvpr} trains a Dust3r-like model for both the local map reconstruction and the global map merging akin to the classic submap-based SLAM system. CUT3R~\cite{wang2025cvpr_cut3r} tries to maintain a token-based scene representation and keeps interacting the scene tokens with the current view tokens for localization and map update. The scene tokens can also be used for novel view synthesis when interacted with a raymap of the novel view. Though not yet popular, implicit token-based scene representation can be used for high-level 3D scene understanding. The integration of these scene tokens into a VLA model would be a promising direction for better handling long-horizon robotics tasks.

\subsection{Global Localization}
Localization refers to estimate 6DOF pose (position and orientation) of robotics. We categorize global localization
 into known map-based localization and relative pose regression based on whether the prior global map is available.
%%%%%%%%需要按照known-map based和无map的对之前的localization方法也举个例子 一两个即可。

\noindent\textbf{\textit{Geometric  Representations}} Traditional known map-based global localization method use point cloud as the 3D scene representation.
If the correspondences (data associations) between the query and the global map are known, the registration problem can be solved in closed form. A representative method is Iterative Closest Point (ICP)~\cite{icp}, which performs point-to-point correspondence search and computes the optimal solution iteratively.  Normal Distributions Transform (NDT)~\cite{ndt} models points as Gaussian distributions and performs registration in a distribution-to-distribution manner. TEASER++~\cite{teaser} introduces a localization approach based on maximum clique, formulating the problem as a graph-theoretic optimization. Voxel-based localization methods~\cite{voxelloc} use voxel-based matching to improve the computational efficiency and real-time performance. Meshloc~\cite{meshloc} introduces a mesh-based localization framework that reduces the overhead of testing local features and feature matchers in visual localization tasks compared to point cloud-based methods. SG-Reg~\cite{sgreg} proposes a scene graph-based point registration network that encodes multiple modalities into semantic nodes and performs coarse-to-fine correspondence matching with a robust pose estimator.

\noindent\textbf{\textit{NeRF-Based Representations}}
iMAP~\cite{imap} is one of the earliest NeRF-Based localization methods, which leverages pixel-level photometric loss to optimize initial rendering poses. LENS~\cite{lens} extends this idea to outdoor environments.  Loc-NeRF~\cite{locnerf} combines traditional Monte Carlo Localization with NeRF-Based mapping, computing particle weights through photometric differences.
Other approaches perform localization in map-free settings by directly estimating relative poses. BARF~\cite{barf} proposes a joint bundle adjustment scheme that simultaneously optimizes NeRF maps and camera poses. Nope-NeRF~\cite{nopenerf} incorporates optical flow and additional constraints by learning undistorted depth maps. Incremental Joint Learning\cite{incremental} further advances this by jointly optimizing depth, pose, and NeRF maps, while adopting a progressive map representation to enable large-scale scene localization.

\noindent\textbf{\textit{3DGS-Based  Representations}}
iComMa~\cite{icomma} is one of the known map-based localization methods. It uses the 2D photometric residual between the query image and the rendered image obtained from a prebuilt Gaussian map to optimize the camera pose. 3DGS-ReLoc~\cite{3dgsreloc} uses a voxel-based 3D Gaussian map with KD-tree structure.
6DGS~\cite{6dgs} introduces a novel ray-casting pipeline combined with an attention-based mechanism that effectively aligns pixel-level image features with 3DGS ellipsoids.
Gsplatloc~\cite{gsplatloc} reconstructs feature-based 3D Gaussian scene representation with robust keypoint descriptor. It uses 2D-3D correspondences between the 3DGS representation and query image descriptors for coarse pose estimation and rendering-based photometric warp loss in the fine stage. For the relative pose estimation, Colmap-free 3DGS~\cite{colmapfree} leverages the temporal continuity from video and
the explicit point cloud representation, achieving accurate view synthesis without SfM pre-processing.
GS-CPR~\cite{GS-CPR} leverages 3DGS as the scene representation and propose a test-time camera pose refinement (CPR) framework. It utilizes a 3D foundation model for 2D matching and enables one-shot pose refinement.
GaussReg~\cite{gaussreg} proposes a coarse-to-fine pose refinement framework. It uses existing point cloud registration methods in coarse stage and uses image-guided registration in fine stage.

\noindent\textbf{\textit{Foundation Model}}
LEXIS~\cite{lexis} explore the use of foundation model CLIP features to perform indoor localization. It builds a real-time topological graph architecture and associate clip feature with graph nodes. FM-Loc~\cite{fmloc} integrates the Large Language Model GPT-3 with the Vision-Language Model CLIP to construct a semantic image descriptor that exhibits robustness to substantial variations in scene geometry and camera viewpoints. In this framework, CLIP is employed to identify objects within an image, GPT-3 subsequently infers candidate room labels based on the detected objects, and CLIP is further utilized to determine the most plausible location label. Anyloc~\cite{anyloc} combine visual foundation model with VLAD descriptor for generalizable visual place recognition and localization without requiring any re-training or fine-tuning. FoundLoc~\cite{foundloc} integrates AnyLoc with a VIO pipeline to achieve visual localization framework. This work is the first to demonstrate that vision foundation model can be effectively deployed on resource-constrained unmanned aerial vehicles (UAVs) and embedded Jetson hardware for state estimation in real-world environments.

% Figures showing applications in medical, large-scale scene, dynamics, and maybe MonoGS

\section{Embodied 3D Interaction}

\label{sec:interaction}
3D interaction is fundamental for robots in real-world environments, encompassing navigation and manipulation. It requires scene representations that capture geometry and semantics in dense, actionable form, enabling robots to understand and interact with their surroundings effectively.

\subsection{Manipulation}
\label{sec:manipulation}
Manipulation poses distinct challenges for scene representation: it requires precise geometry, semantic grounding, and real-time decision support for object-level reasoning and user-guided actions. Beyond single-format encodings, many recent approaches adopt hybrid models---combining, for example, neural fields with voxel maps or enriching geometry with language-conditioned tokens---reflecting a shift toward task-adaptive 
and integrated representations.

\noindent\textbf{\textit{Geometric Representations}}
Traditional manipulation pipelines often rely on explicit geometric encodings such as points, voxels, or signed distance fields. Point-based approaches directly operate on raw clouds, enabling lightweight and real-time grasp prediction; for example, Contact-GraspNet~\cite{sundermeyer2021contact} and PointNetGPD~\cite{qi2019pointnetgpd} infer 6-DoF grasps from point sets via affordance cues. 
Voxel grids discretize 3D space to capture dense occupancy or features, and have been widely applied in grasp detection and manipulation.  
Representative works include SCG~\cite{varley2017shape}, which voxelizes local point cloud regions to evaluate grasp quality,  
VoxelNet-based grasp pose estimation~\cite{zhou2018voxelnet}, which applies 3D CNNs over voxelized inputs,  
and Volumetric Grasping Network (VGN)~\cite{breyer2021volumetric}, which directly predicts 6-DoF grasp distributions from TSDF voxel grids. RoboEXP~\cite{roboexp} and DovSG~\cite{dovsg} leverages scene graph representation for interactive exploration and long-horizon, language-guided task planning.

% Voxel grids discretize 3D space to capture dense occupancy or features, with methods such as NDF~\cite{simeonov2022neural}, Local-NDF~\cite{chun2023local}, and GIGA~\cite{jiang2021synergies} modeling graspability in volumetric form, while VoxPoser~\cite{huang2023voxposer} integrates voxel maps with language models for spatial goal planning. Signed Distance Fields (SDFs) provide smooth implicit surfaces and accurate contact estimation, as in VGN~\cite{breyer2021volumetric}, NGDF~\cite{weng2022ngdf}, and NeuralGrasp~\cite{khargonkar2023neuralgrasps}. 
While effective, these geometric encodings face common limitations: point clouds are sparse and occlusion-prone, voxel grids are memory-intensive and resolution-limited, and SDFs require volumetric storage and costly surface extraction, motivating the shift toward neural and hybrid representations.

\noindent\textbf{\textit{NeRF-Based Representations}}
NeRF-Based approaches reconstruct appearance and geometry for photorealistic manipulation and teleoperation. Dex-NeRF~\cite{ichnowski2021dexnerf}, GraspNeRF~\cite{Dai2023GraspNeRF}, Evo-NeRF~\cite{kerr2022evo}, and Radiance Fields for Robotic Teleoperation~\cite{patil2024radiance} provide scene priors for interactive control. However, vanilla NeRFs suffer from slow rendering and lack of editability, which motivates structured extensions.

\noindent\textbf{\textit{3DGS-Based Representations}}
3DGS offers an explicit, differentiable representation with fast rasterization and geometric flexibility. GaussianGrasper~\cite{zheng2024gaussiangrasper}, ManiGaussians~\cite{lu2024manigaussian}, GraspSplats~\cite{ji2024graspsplats}, and Point'n Move~\cite{huang2024point} demonstrate manipulation control via dynamic Gaussians. HGS-Planner~\cite{xu2025hgs}, Object-Aware GS~\cite{li2024objectaware}, MANUS~\cite{pokhariya2024manus}, and Physically Embodied GS~\cite{abou2024physically} show applications in reconstruction, modular control, articulated interaction, and feedback-based correction. SplatSim~\cite{qureshi2024splatsim} and RL with Generalizable GS~\cite{wang2024reinforcement} further support sim2real transfer. These are promising for interactive training and low-latency policy learning.

\noindent\textbf{\textit{Foundation Model}}
Token-based scene representations bridge LLMs/VLMs and 3D environments. Feature fields like DFF~\cite{shen2023distilled}, LERF~\cite{kerr2023lerf}, and GNFactor~\cite{Ze2023gnfactor} tokenize space via language-conditioned features. Ditto~\cite{jiang2022ditto} constructs digital twins through interaction, while DreamHOI~\cite{zhu2024dreamhoi} and EnerVerse~\cite{huang2025enerverse} use diffusion priors for interaction modeling. NLMap~\cite{chen2023open} enables spatial language grounding through a semantic point-based representation, supporting open-vocabulary queries for real-world planning. Though not token-based, SayCan~\cite{ichter2022saycan} and Code as Policies~\cite{codeaspolicies2022} optionally incorporate tokenized or voxelized modules for spatial reasoning and language-conditioned planning.

\subsection{Navigation}
\begin{figure}[t]
    \centering
    \includegraphics[width=\linewidth]{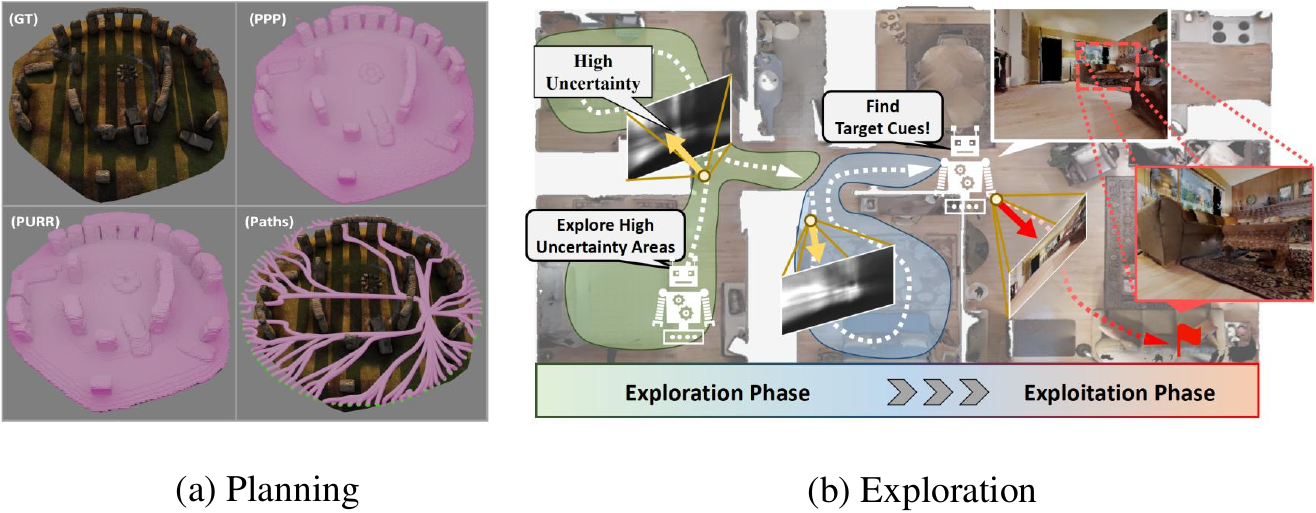}
    \vspace{-0.5cm}
    \caption{ An illustration of navigation methods based on neural scene representation. Subplot(a) to subplot(b) are extracted from \cite{catnips},~\cite{nerfnav}.
}
\vspace{-0.5cm}
    \label{fig:nav}
\end{figure}
Robotic navigation requires accurate perception and scene representation of their environments in order to plan collision-free trajectories. We summarize the key modules of 3DGS-Based navigation, including path planning, and exploration, shown in Fig.~\ref{fig:nav}.

% \begin{table}[!t]
% \centering
%   \scalebox{0.95}{
%   \setlength{\tabcolsep}{1.2mm}{
% \begin{tabular}{lccc}
% \toprule

% \textbf{Methods} & \textbf{Acc. (cm) ↓} & \textbf{Comp. (cm) ↓} & \textbf{Comp. Ratio (\%) ↑} \\
% \midrule
% FBE \cite{fbe}         & /     & 9.78 & 71.18 \\
% UPEN \cite{upen}       & /     & 10.60 & 69.06 \\
% OccAnt \cite{occant}   & /     & 9.40 & 71.72 \\
% ANM~\cite{anm}       & 7.80 & 9.11 & 73.15 \\
% NARUTO~\cite{naruto}  & 6.31 & 3.00 & 90.18 \\
% ActiveGamer~\cite{activegamer} & \textbf{1.66} & \textbf{2.30} & \textbf{95.32} \\
% \bottomrule
% \end{tabular}}}
% \caption{Comparison of different methods on Matterport3D dataset~\cite{matterport3d}.}
% \label{tab:comparison}
% \end{table}

\noindent\textbf{\textit{Geometric Representations}}
Traditional navigation methods~\cite{pointnav,pointnav2} leverages point cloud-based representation for real-time navigation. OctoMap~\cite{octomap} provides a volumetric representation of space using octrees with probabilistic occupancy estimation, explicitly modeling not only occupied regions but also free and unknown areas. It further introduces an octree-based map compression strategy to maintain compact 3D models. The Fast-Planner~\cite{fastplanner} series of works leverage voxel maps to develop a robust and efficient quadrotor motion planning system, enabling agile flight in complex three-dimensional environments. Other approaches employ SDF representations for planning. Voxblox~\cite{voxblox} incrementally builds an Euclidean Signed Distance Field (ESDF) from a Truncated Signed Distance Field (TSDF), which directly supports trajectory optimization and runs entirely on-board in real time. FIESTA~\cite{fiesta} proposes a gradient-based ESDF representation, where two independent updating queues are designed for inserting and deleting obstacles separately. 

\noindent\textbf{\textit{NeRF-Based Representations}}
NeRF-Navigation~\cite{nerfnavigation} enables safe navigation using a NeRF map by penalizing potential collisions between the robot body's point-cloud model and the neural field. NFOMP~\cite{nfomp} learns an obstacle neural field to facilitate obstacle avoidance while performing online trajectory optimization. CATNIPS~\cite{catnips} reinterprets the density field as a collection of points in continuous space following a Poisson Point Process, which allows for a principled quantification of collision probability. RNR-Map~\cite{rnrmap} further proposes a visual navigation framework using renderable neural radiance map with latent codes at each pixel. \cite{nerfnav} proposes a novel navigation pipeline based on NeRF map with Uncertainty-driven exploration. It uses the memory information from the NeRF map to strengthen the robot's reasoning capabilities for determining target locations. 

\noindent\textbf{\textit{3DGS-Based Representations}}
 Some methods use 3DGS-Based map to generate a path to the goal, achieving promising performance. GaussNav~\cite{gaussnav} proposes a novel framework for Instance ImageGoal Navigation, which enables the agent to effectively encode scene geometry and semantics while preserving object-level textural features through the 3DGS representation. This method constructs a semantic Gaussian map and projects it onto 2D BEV grids. By matching the rendered appearances of similar objects with the given target, the agent can accurately identify, localize, and navigate to the specified object. Splat-Nav~\cite{splatnav} designs two components: a safe planning module, and a robust vision-based pose estimation module. The plan module builds a safe-by-construction polytope corridor through the map based on
mathematically rigorous collision constraints and then constructs
a Bézier curve trajectory through this corridor.  
BEINGS~\cite{beings} proposes a Bayesian Embodied Image-goal Navigation framework with 3DGS scene representation. It reduces data dependency and allowing dynamic adjustment of ImageNav strategies using Bayesian updates informed by real-time data. Some methods use 3DGS scene representation for exploration an unobserved map. GS-planner~\cite{gsplanner} designs an active  high-fidelity reconstruction using 3D Gaussian Splatting. It evaluates the reconstruction quality and completeness of 3DGS map online to guide the robot exploration.
Beyond Uncertainty~\cite{beyondnav} proposes the Risk-aware Environment Masking (RaEM) framework. It leverages coherent risk measures to dynamically prioritize safety-critical regions of the unknown environment, guiding active view acquisition algorithms toward identifying the next-best-view (NBV).

\noindent\textbf{\textit{Foundation Model}}
Compared to traditional representations and NeRF/3DGS-Based models, the most straightforward application of Vision-Language Models (VLMs) in robotics is to leverage their ability to perform open-set object recognition and scene understanding for navigation. NLMap~\cite{nlmap} constructs an open-set and queryable scene representation using a large language model (LLM) to ground task plans, enabling language-based goal-oriented planning. LM-Nav~\cite{lm-nav} utilizes an LLM to extract landmarks for navigation from natural language instructions, which are then grounded in a pre-built graph via a VLM. A planning module is subsequently employed to guide the robot to the specified landmarks. VLN-BERT~\cite{vln-bert} introduces a visual-linguistic transformer model that combines multi-modal visual and language representations for visual navigation, using web-based data.

Additionally, foundation models enable task-level planning, where complex tasks are broken down into smaller actionable steps. SayCan~\cite{saycan} serves as a representative example of task-level planning, where LLMs are used for high-level task planning. Similarly, VLP~\cite{vlp} enhances long-horizon planning by integrating a text-to-video dynamics model. These task-level methods do not require precise execution of sub-tasks in the environment, as they rely on pre-defined or pre-trained skills. Reasoned Explorer~\cite{explorerreasoning} uses LLMs as evaluators to score nodes in a 2D undirected graph, which serves as a map for both visited points and frontier evaluations. This approach couples with external memory and incremental map-building, effectively addresses the context length limitation of LLMs. VoxPoser~\cite{voxposer} applies VLMs to obtain affordance functions, referred to as 3D value maps, for motion planning. Saytap~\cite{saytap} introduces the innovative concept of utilizing foot contact patterns as action representations, where the language model outputs a `0' for no contact and a `1' for floor contact, allowing LLMs to generate zero-shot actionable commands for quadrupedal locomotion tasks such as jumping and jogging.

\section{Discussions and Future Directions}
In this section, we discuss challenges related to 3D scene representation in robotics. We also explore
potential future avenues to address some of these challenges.
% 感知规划一体式场景表示：BEV
\subsection{General-purpose or Module Specify}
At present, most robotic systems are built upon modular intelligence, where functions such as navigation or manipulation are decomposed into separate modules, such as perception, mapping, localization, manipulation, and navigation to accomplish complex tasks. This design facilitates the implementation of diverse robotic capabilities. However, such modularity may inherently constrain the advancement of robotics intelligence. While modular solutions introduce useful inductive biases and support effective task-specific performance, they often suffer from limited generality and poor transferability. In practice, they usually require repeated sensor calibration, environment-specific modeling, and parameter re-tuning across different scenarios. Moreover, in highly complex environments, constructing accurate models remains particularly challenging. Recent advances in foundation models offer an alternative pathway by enabling end-to-end intelligence.

 Drawing an analogy to the human brain, neuroscientists have identified specialized regions such as the visual cortex, somatosensory cortex, and motor cortex. Yet, the brain exhibits remarkable plasticity and the capacity to reorganize its functions to adapt to environmental changes or neural damage. Prior work~\cite{dynamicbrain} highlights that during the learning of human brain, the brain possesses both the flexibility to adapt existing functions and the precision to recruit new neurophysiological activities to drive desired behaviors. These two attributes---flexibility and selection---operate across multiple temporal scales as skills evolve from being effortful and slow to automatic and efficient. This flexibility suggests that modularity in the brain may have emerged as a byproduct of unified training rather than as an inherently isolated design. Inspired by this perspective, the pursuit of general-purpose robotics calls for foundation models that leverage expansive datasets and large-scale deep learning architectures to integrate perception, reasoning, and action in a unified manner. Similarly, in Bertology~\cite{bertology}, researchers have shown that local components of large-scale models can gradually specialize in specific functions through continued training.
 
\subsection{Data scarcity with generative model}
Despite the advantages of neural scene representations in terms of accuracy and generality, a major challenge lies in the scarcity of robotics-specific data compared to the internet-scale text and image corpora used to train LLMs and VLMs. This limitation significantly hinders the development of neural scene representations and foundation models for robotics. To address this issue, research has focused on enhancing the generalization ability of neural scene representations under limited data regimes. One line of work, ROSIE~\cite{rosie} explores data augmentation through generative models or simulators, aiming to create semantically and visually diverse data while ensuring physical feasibility and accuracy. Another direction leverages world models to predict state transitions conditioned on actions, thereby generating additional training datasets. For example, the Navigation World Model (NWM)~\cite{nwm} is a controllable video generation model that predicts future visual observations from past egocentric views and navigation behaviors. NWM employs a Conditional Diffusion Transformer (CDiT) trained on a diverse collection of human and robotic egocentric videos, enabling trajectory planning by simulating and evaluating whether candidate paths achieve the target goal. Unlike supervised navigation policies with fixed behaviors, NWM can flexibly integrate constraints during planning. On the data side, large-scale real-world collection efforts remain essential, such as Openx-embodiment~\cite{Openx-embodiment}, but simulation platforms such as Habitat provide a complementary solution. Habitat-Sim~\cite{habitat} can render thousands of frames per second, allowing the construction of large-scale, high-fidelity datasets for robot learning.
% 长期任务的场景表示
\subsection{Real-time Performance}
Compared with traditional scene representations, another critical bottleneck for deploying neural scene representations in robotics lies in their inference time, which remains a limiting factor for reliable real-time applications. Current deployment strategies of neural networks generally fall into two categories. The first is cloud-based deployment, where large foundation models such as DINOv2~\cite{dinov2}, SAM~\cite{sam}, and GPT~\cite{gpt4} are typically hosted in remote data centers and accessed via APIs. In this paradigm, response latency and service time depend heavily on underlying network routing, bandwidth, and data center computation. Thus, network reliability and latency must be carefully considered before integrating such models into autonomous robotic stacks. The second approach is onboard deployment on edge-computing platforms, where techniques such as model distillation and quantization~\cite{awq} are employed to reduce model size and enable real-time inference. However, this often comes at the cost of reduced generalization capability. A promising future direction lies in hardware--algorithm co-design, aiming to simultaneously improve inference efficiency and maintain model generalization for real-time robotic deployment.

\section{Conclusion}
We reviewed recent literature on scene representations in robotics, comparing their strengths and limitations across perception, mapping, localization, manipulation, and navigation. We also discussed future directions, questioning whether robotics should retain modular architectures or move toward unified, foundation model-based approaches. Key opportunities include generalization, zero-shot capabilities, human-robot interaction, and scalability, all of which could transform robotics. Yet significant challenges remain for real-world deployment, and we highlight the obstacles and research avenues needed to overcome them.
% Through a review of the recent literature, we have explored the diverse and promising applications of various scene representations in robotics. We compared the strengths and limitations of different scene representations, investigating their effectiveness across multiple robotic modules such as perception, mapping, localization, manipulation, and navigation. Furthermore, we have discussed the future directions of scene representation for real-world robotic applications, addressing the question of what paradigms are needed moving forward. Specifically, we consider whether the modular architecture currently used in robotics should be maintained or if a more unified, foundation model-based general-purpose approach is preferable. We believe that generalization, zero-shot capabilities, advanced human-robot interaction, and scalability hold the potential to significantly transform the field of robotics. However, current methods still face challenges in achieving real-world deployment. In our discussion, we recognize the obstacles and potential risks that must be tackled in future research and have outlined possible avenues for improvement.

\tiny
    \bibliographystyle{IEEEtran}
    \bibliography{main}

\vfill

\end{document}